\documentclass[10pt,twocolumn,letterpaper]{article}

\usepackage[dvipsnames,svgnames,table]{xcolor}

\usepackage[pagenumbers]{cvpr} 

\usepackage[dvipsnames]{xcolor}

\newcommand{\cm}{{\ding{51}}}
\newcommand{\xm}{{\ding{55}}}

\usepackage{times}
\usepackage{epsfig}
\usepackage{graphicx}
\usepackage{amsmath}
\usepackage{amssymb}
\usepackage{bm}
\usepackage{graphbox}

\usepackage{makecell}
\usepackage{booktabs}
\usepackage{tabularx}
\usepackage{multirow}
\usepackage{pifont}
\usepackage{tablefootnote}
\usepackage[para,online,flushleft]{threeparttable}

\usepackage{algorithm}
\usepackage{setspace}
\usepackage[noend]{algpseudocode}

\usepackage{color}

\usepackage[accsupp]{axessibility}

\definecolor{cvprblue}{rgb}{0.21,0.49,0.74}
\usepackage[pagebackref,breaklinks,colorlinks,citecolor=cvprblue,hypertexnames=false]{hyperref}

\def\paperID{8203}
\def\confName{CVPR}
\def\confYear{2024}

\title{AdaBM: On-the-Fly Adaptive Bit Mapping for Image Super-Resolution}

\author{
Cheeun Hong$^1$ \hspace{3cm} Kyoung Mu Lee$^{1,2}$ \\
{$^1$ Dept. of ECE \& ASRI, $^2$ IPAI, Seoul National University, Seoul, Korea} \\
{\tt\small \{cheeun914, kyoungmu\}@snu.ac.kr} \\
}

\begin{document}

\maketitle
\begin{abstract}
Although image super-resolution (SR) problem has experienced unprecedented restoration accuracy with deep neural networks, it has yet limited versatile applications due to the substantial computational costs.
Since different input images for SR face different restoration difficulties, 
adapting computational costs based on the input image, referred to as adaptive inference, has emerged as a promising solution to compress SR networks.
Specifically, adapting the quantization bit-widths has successfully reduced the inference and memory cost without sacrificing the accuracy.
However, despite the benefits of the resultant adaptive network, existing works rely on time-intensive quantization-aware training with full access to the original training pairs to learn the appropriate bit allocation policies, which limits its ubiquitous usage.
To this end, we introduce the first on-the-fly adaptive quantization framework that accelerates the processing time from hours to seconds.
We formulate the bit allocation problem with only two bit mapping modules: one to map the input image to the image-wise bit adaptation factor and one to obtain the layer-wise adaptation factors.
These bit mappings are calibrated and fine-tuned using only a small number of calibration images.
We achieve competitive performance with the previous adaptive quantization methods, while the processing time is accelerated by $\times$2000.
Codes are available at \href{https://github.com/Cheeun/AdaBM}{https://github.com/Cheeun/AdaBM}.
\end{abstract}

\section{Introduction}
\label{sec:intro}
Image super-resolution (SR) is a classic computer vision problem that aims to restore the high-resolution image (HR) from the corresponding low-resolution input image (LR).
Since the emergence of deep neural networks, SR has been able to produce high-resolution, high-fidelity outputs.
However, achieving such a level of quality was conditional upon utilizing computationally heavy SR models, necessitating demanding computing power and storage costs.
Moreover, as the industry's demand is stepping towards super-resolving larger inputs (\eg, 4K TVs), the computational burden has increased quadratically.
Therefore, recent interest has shifted to reducing the computational costs of SR networks with little or no sacrifice on restoration accuracy.
Among the surge towards lightweight SR models, quantization is one promising avenue to reduce the memory/inference cost of neural networks by replacing the 32-bit floating-point (FP) weights and activations with lower precision values.

\begin{table}[t]
    \centering
    \resizebox{0.47\textwidth}{!}{
        \begin{tabular}{l cccc rr}
            \toprule
            Method & Adaptive & Data & GT & BS & Iterations & Process Time\\
            \midrule
            EDSR~\cite{lim2017enhanced} & - & 800 & \cm & 16 & 150,000 & 30 hrs\\
            \midrule
            PAMS~\cite{Li2020pams} & \xm & 800 & \cm & 16 & 15,000 & 2 hrs\\
            DDTB~\cite{zhong2022ddtb} & \xm & 800 & \cm & 16 & 30,000 & 4 hrs \\
            PTQ4SR~\cite{tu2023toward} & \xm & 100 & \xm & 2 & 500 & {124 sec} \\
            \midrule
            CADyQ~\cite{hong2022cadyq} & \cm & 800 & \cm & 16 & 200,000 & 40 hrs\\
            CABM~\cite{tian2023cabm} & \cm & 800 & \cm & 16 & 350,000 & 70 hrs\\
            \textbf{AdaBM (Ours)} & \cm & 100 & \xm & 2 & 500 & \textbf{71 sec}\\
            \bottomrule
        \end{tabular}
    }
    \vspace{-0.2cm}
    \caption{ 
    \textbf{Existing methods for quantization on SR.}
    Adaptive denotes whether bit-widths are adaptively allocated, GT denotes the requirement for ground-truth HR images, and BS denotes the batch size.
    Metrics are reported for quantizing EDSR ($\times$4).
    }
    \label{tab:intro-first}
\end{table}

\begin{figure*}[t]  
    \centering
    \includegraphics[width=0.95\linewidth]{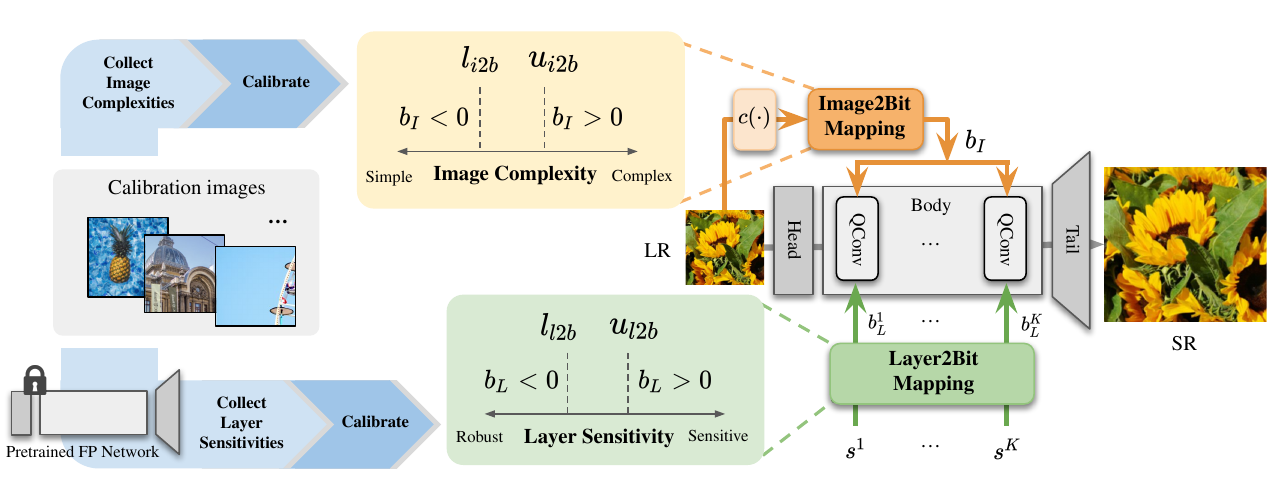}
    \vspace{-0.4cm}
    \caption{
        \textbf{Illustration of our adaptive bit-mapping.} 
        During inference, an input image is mapped to the image-wise bit adaptation factor based on its complexity (Image2Bit Mapping),
        then together with the layer-wise bit adaptation factors pre-determined based on layer-wise sensitivities (Layer2Bit Mapping), the two factors adapt the quantization bit-widths.
        Higher bit-widths are assigned to sensitive layers and complex images.
        The thresholds of mapping modules are calibrated and fine-tuned using a small set of calibration images. 
    }
    \label{fig:method-overview}
\end{figure*}

Quantizing SR networks with minimal accuracy loss has been a challenging problem, as SR network activations exhibit predominantly variant distributions during test time, often leading to severe quantization errors.
Few works address variant activations using a more fine-grained quantization function~\cite{hong2022daq} or updating the quantization ranges to better fit dynamic variant activations~\cite{Li2020pams,zhong2022ddtb}.
However, these methods neglect that the accuracy loss from quantization differs for different images and layers of the network.
Some images are easier to reconstruct than others; in other words, they can still be accurately reconstructed with fewer computations.
Thus, assigning lower bit-widths to such images leads to a better computational cost and accuracy trade-off.
In this spirit, recent methods~\cite{hong2022cadyq,tian2023cabm} take into account the different quantization sensitivity of the images and adapt the bit-widths based on the input content.

To adaptively allocate bit-widths to images, existing methods~\cite{hong2022cadyq,tian2023cabm} employ several quantization function candidates of different bit-widths for each activation.
Then, CADyQ~\cite{hong2022cadyq} trains MLPs that predict the probability for each quantization function, in which the function with the highest probability is selected at the test time.
Additionally, CABM~\cite{tian2023cabm} builds a lookup table based on the trained MLPs to find a better bit-width for the input image.
Although previous methods have achieved quantization on SR models with minimal accuracy loss, these approaches involve extensive quantization-aware training (QAT) with the full training dataset of LR and HR pairs.
For example, as reported in Table~\ref{tab:intro-first}, it takes 40/70 hours to obtain the final quantized model for CADyQ and CABM, respectively.
In short, searching the proper bit-widths for each layer and image can provide a better accuracy-complexity trade-off, but the search cost for bit assignment is substantial.

To this end, we propose the first adaptive bit-mapping framework for image super-resolution that adapts the bit-widths for different images \textbf{on-the-fly}.
Based on our observation that image-wise variance of quantization error and layer-wise variance are independent, we find that image and layer-wise adaptation can be separately processed.
This dramatically reduces the search cost as only two policies are required for bit allocation: one to determine the image-wise adaptation factor for the test images and another to determine the layer-wise adaptation factors for all layers.
Also, since we fix the layer-wise adaptation to be invariant of the input image, layer-wise adaptation factors can be pre-determined before test time.
This allows our bit-mapping to be learned within the second level using a small subset of calibration LR images without corresponding HR.

For image-wise bit adaptation, we design an image-to-bit mapping module that maps an image to an image-wise bit adaptation factor based on the complexity of the image.
The adaptation factor for an image is obtained using complexity thresholds; images that are more complex than the upper complexity threshold are mapped to a positive adaptation factor that adjusts the bit-width to a higher bit.
Then, calibration images are used to calibrate and fine-tune the complexity thresholds.
Similarly, layer-wise bit adaptation factors are determined by the layer-to-bit mapping module based on each layer's sensitivity to quantization.
The sensitivity of the layer is calculated by processing the calibration images with the pre-trained FP network.
The sensitive layers (larger than the upper sensitivity threshold) are mapped to a positive adaptation factor.
Layer-wise adaptation factors are calibrated and directly fine-tuned using calibration images.
Furthermore, to find a better quantization range for each layer-wise variant bit-width, we clip the range by optimizing the L2 distance between the quantized tensor using the assigned layer-wise bit and the FP counterpart.

In summary, we accelerate the expensive training process for adaptive quantization by calibrating the quantization parameters (ranges and bit-mappings) and then fine-tuning the quantization parameters for a few iterations with the calibration images.
We achieve on-par performance with QAT-based methods with $\times2000$ less processing time, pushing adaptive quantization to a new frontier.
\section{Related works}
\label{sec:related_works}
\begin{figure*}[t]  
    \centering
    \newcommand{\w}{0.33\linewidth}
    \begin{subfigure}{\w}
        \centering
        \includegraphics[height=0.8\linewidth]{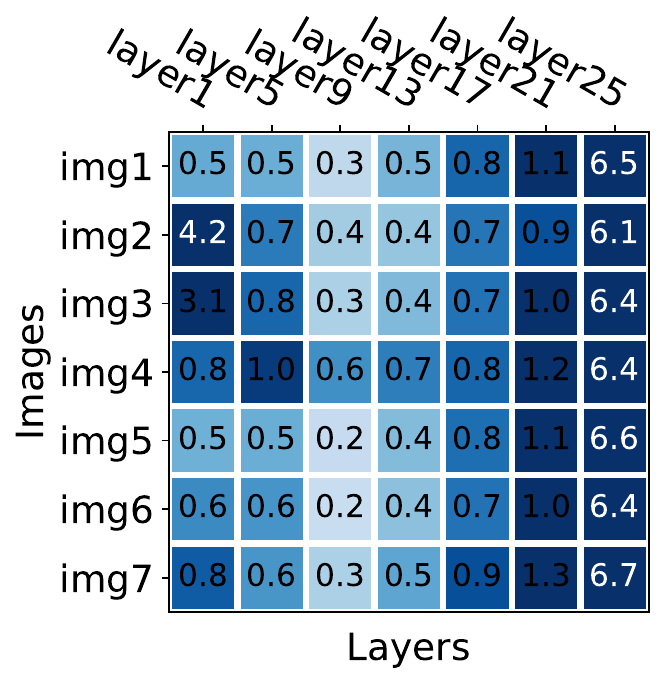}
        \caption{
            Image and layer-wise MSE
            \label{fig:method-obs-a}
        }
    \end{subfigure}
    \begin{subfigure}{\w}
        \centering
        \includegraphics[height=0.77\linewidth]{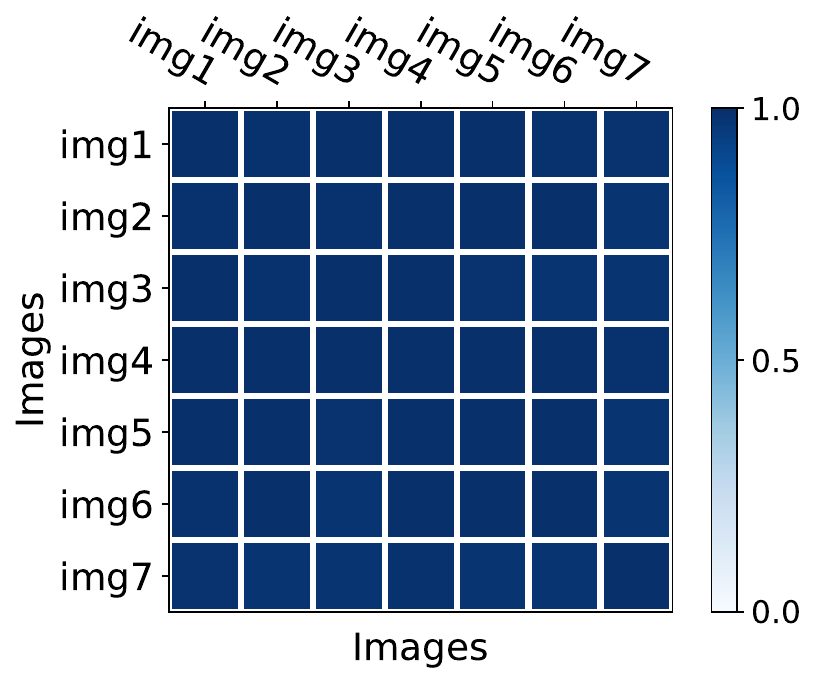}
        \caption{
            Cosine similarity between layer-wise MSE
            \label{fig:method-obs-b}
        }
    \end{subfigure}
    \begin{subfigure}{\w}
        \centering
        \includegraphics[height=0.8\linewidth]{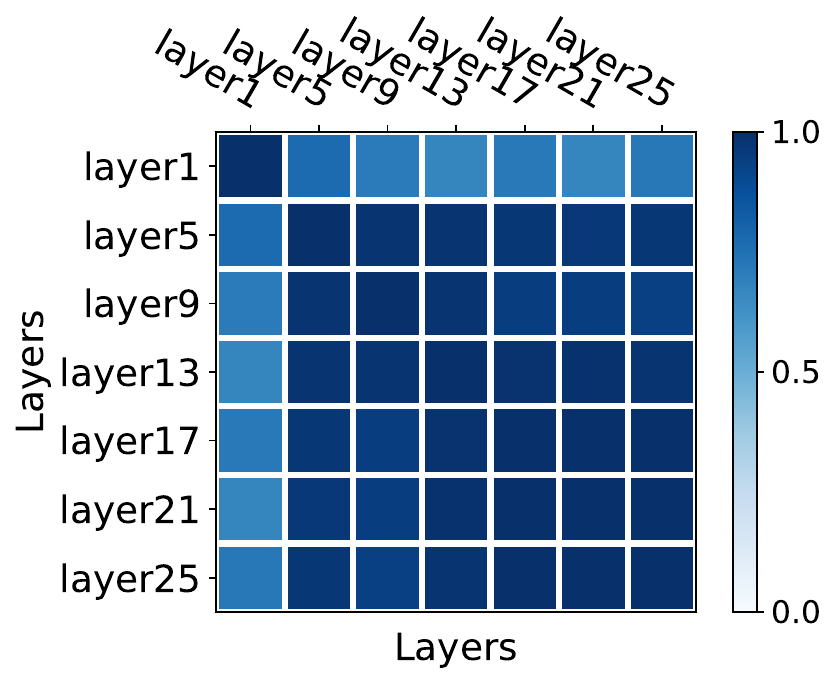}
        \caption{
            Cosine similarity between image-wise MSE
            \label{fig:method-obs-c}
        }
    \end{subfigure}
    \vspace{-0.6cm}
    \caption{
        \textbf{Analysis on layer-wise and image-wise quantization sensitivity.} 
        (a) Mean squared error (MSE) between statically quantized activation and corresponding FP is different per layer and per input image. 
        (b) The \textit{relative} orders of layer-wise MSEs tend to be invariant of the input image, as layer-wise MSEs of different images have high cosine similarity values.
        (c) Similarly, the \textit{relative} orders of image-wise MSEs are consistent throughout network layers.
        This indicates that layer-wise bit adaptation and image-wise bit adaptation can be done separately, which effectively reduces the search cost for bit allocation.
    }
    \label{fig:method-obs}
\end{figure*}

\noindent
\textbf{Image super-resolution models.}
Deep convolutional neural networks advanced the performance of image super-resolution (SR) problem~\cite{lim2017enhanced,zhang2018residual,ledig2017photo,ahn2018fast,Park_2023_ICCV}.
However, these advances rely on enormous computing power and storage costs, which limits the applicability of these networks, such as deployment to mobile devices.
Such limitations initiated the prevailing research for lightweight SR networks.
New lightweight models are designed manually~\cite{dong2016srcnn,hui2019imdn,hui2018idn,song2021addersr,ma2019efficient,xin2020binarized} or searched~\cite{chu2021fast,kim2022fine,li2020dhp,li2021heterogeneity}.
Additionally, various compression techniques are investigated to compress existing pre-trained SR networks, such as quantization~\cite{Li2020pams,Wang2021fully,ayazoglu2021extremely,hong2022daq,zhong2022ddtb,qin2023quantsr}, pruning~\cite{zhan2021achieving,jiang2021learning,zhang2021learning,Oh_2022_CVPR}, and knowledge distillation~\cite{zhang2021data,gao2018image}.

\noindent
\textbf{Quantizing image super-resolution models.}
Due to the benefits of cut-down memory and inference cost, quantization, which maps 32-bit floating point values of weights and activations to lower-bit values, became a promising solution for heavy SR models.
Quantizing SR models with minimal accuracy degradation is challenging, since most SR networks~\cite{lim2017enhanced,zhang2018residual} have vastly distinct activation distributions.
Consequently, researchers utilize learnable quantization range for different layers~\cite{Li2020pams,qin2023quantsr,hong2023overcoming}, adopt channel-wise quantization function~\cite{hong2022daq}, or design a dynamic module to adapt quantization range during test-time~\cite{zhong2022ddtb}.
Although these works apply a fixed quantization level throughout the SR network, a work manually assigns different bit-width for each stage of the SR network~\cite{liu2021super}.
Furthermore, recent works~\cite{hong2022cadyq,tian2023cabm} design an adaptive quantization framework with a bit-width prediction module that dynamically adapts the quantization level according to the content of the input image.

\noindent
\textbf{Adaptive inference.}
To make deep learning practical, allowing a pre-trained model to be responsive and adaptable to various deployment scenarios and resource constraints is crucial.
Several works introduced adaptive SR networks to achieve efficient inference for a given input image~\cite{kong2021classsr,yu2021path,wang2022adaptive,hong2022cadyq,tian2023cabm,liu2020deep,wang2021exploring,xie2021fadn}.
These works either adaptively reduce the network depth~\cite{yu2021path,wang2022adaptive} or the number of channels~\cite{kong2021classsr}, or assign adaptive bit-widths to replace floating-point operations~\cite{hong2022cadyq,tian2023cabm}.
However, although adaptive inference enables improved performance-complexity trade-off by assigning more computations to inputs more in need, such trade-off comes at the cost of heavy re-training with the full training dataset of LR, HR pairs.

\section{Proposed method}
\label{sec:proposed_method}

\subsection{Preliminaries}
As SR networks are known to exhibit highly asymmetric activation distributions~\cite{zhong2022ddtb,hong2022daq,hong2022cadyq,tian2023cabm}, we follow the common practice of employing an asymmetric uniform quantizer for activations.
The asymmetric quantizer is parameterized by the lower and upper bounds $l, u$ and the bit-width $b$, in which the quantized integer is produced as follows:
\begin{equation}
    \bm{X}_q =q(\bm{X};b,l,u) = \lfloor\frac{clip(\bm{X},l,u)}{S}-l\rceil \cdot S+l,
    \label{eq:quantizer}
\end{equation}
where $clip(\cdot,l,u)=min(max(\cdot,l),u)$, $\lfloor\cdot\rceil$ rounds the tensor to the nearest integer, and scaling factor $S = \frac{u-l}{2^b -1}$ converts the range of the floating-point values to the range of $b$-bit.
For weights, we employ a symmetric quantizer.

\subsection{Motivation}
To obtain an accurate quantizer, finding a proper clipping range for activations $[l_a,u_a]$ is important~\cite{Li2020pams,zhong2022ddtb,qin2023quantsr}, but the bit-width $b$ that determines the quantization level is also a crucial factor.
Adopting higher bit-widths for quantization-sensitive layers and lower bit-widths for quantization-robust layers has been proven effective~\cite{dong2019hawq,wang2019haq,chen2021towards,tang2022mixed}.
Especially on SR networks, as quantization sensitivity largely varies for different layers and input images, adapting the bit-width of all quantizers based on the complexity of the input image and the sensitivity of the layer-wise input activations can provide better accuracy~\cite{hong2022cadyq,tian2023cabm}.
However, finding a proper policy for such bit-width adaptation is time-consuming; for $K$ convolutional layers and $M$ bit-width candidates, the network has to learn $K\cdot M$ policies to predict the bit-width.
In addition, inference with $K$ adaptive modules incurs additional inference costs.

In contrast, we observe that the layer-wise and image-wise adaptation can be processed separately, serving as a key to avoiding the extensive bit-allocation policy search.
First, we observe that the \textit{relative} order of layer-wise sensitivities tends to be consistent regardless of the input image.
As shown in \Cref{fig:method-obs}, the layer-wise quantization errors of different images have high cosine similarity values, indicating that relative layer-wise sensitivities are independent of the image.
Moreover, we find that the \textit{relative} image-wise variant sensitivities of a specific layer are preserved throughout network layers.
This means that layer-wise adaptation can be learned relatively among the layers and image-wise adaptation can be learned relatively within images.
This simplifies the bit allocation problem to two policies: one policy for image-wise adaptation and one policy for layer-wise adaptation, which facilitates quick learning of bit-width allocation.
Also, given that the layer-wise adaptation policy is invariant to the input image, it can be pre-determined before test time, thus only one adaptive module is processed during inference.
In short, the static bit-width for the quantizing tensor of the $j$-th image of the mini-batch and the $k$-th convolutional layer, $b^{j,k}$ is adapted separately by the image-wise adaptation factor $b_I^j$ and layer-wise adaptation factor $b_L^k$ as follows:
\begin{equation}
    b^{j, k} = b_{base} + b_I^j + b_L^k,
    \label{eq:bit}
\end{equation}
The following sections describe how each factor is obtained.

\subsection{Complexity-based image-to-bit mapping}
Our goal is to determine a single universal bit adaptation factor for image-wise adaptation of the whole quantized network.
This factor adaptively decreases or increases the layer-wise bit-widths based on the input image.
To determine the bit-width factor for each image, image complexity can be an effective guiding metric.
Image complexity can hint at quantization sensitivity; it is known that complex images (\ie, larger edge density, higher spatial frequency, larger color diversity) tend to require more computational costs of reconstruction. 
Therefore, we design a mapping module that directly maps an image to a bit-width factor based on the complexity of the image.
When the complexity of an image is larger than the complexity upper bound, it is mapped to a positive bit-width factor that globally increases the bit-widths of the network at the test time.
Similarly, during inference, an image of complexity smaller than the complexity lower bound is mapped to a negative factor that globally decreases layer-wise bit-widths. 
The image-to-bit mapping function that maps the complexity of the $j$-th image of the mini-batch $I_{LR}^j$ to bit adaptation factor $b_I^j$ is formulated as follows:
\begin{equation}
    b_I^j = I2B(c(I_{LR}^j)) =
    \begin{cases}
    -1, & c(I_{LR}^j) < l_{i2b}, \\
    +1, & c(I_{LR}^j) > u_{i2b}, \\
    0, & otherwise,
    \end{cases}
    \label{eq:i2b}
\end{equation} 
where $c(\cdot)$ measures the complexity of the input image by calculating the average gradient density~\cite{hong2022cadyq} of the image. 
We note that we simply set the bit factor to be $\{-1,0,1\}$, but it can be modified based on the hardware capacity.
To determine the complexity thresholds of the image-to-bit mapping module $l_{i2b}, u_{i2b}$, we collect the complexity measure of the small set of calibration images.
Then we use the $p_I$-th and $(100-p_I)$-th percentiles of the complexity measures obtained to initialize $l_{i2b}$ and $u_{i2b}$, respectively.
Although the statistics serve as decent thresholds for allocating image-adaptive bit-width factors, we are able to obtain better sweet spots by fine-tuning the thresholds with the calibration images.

\subsection{Sensitivity-based layer-to-bit mapping}
It is commonly recognized that different layers of the network have different sensitivity to quantization~\cite{dong2019hawq,wang2019haq,chen2021towards,tang2022mixed}.
Certain layers are more robust to quantization than others, thus allocating low bit-widths to these layers can lead to a better trade-off between computational complexity and accuracy.
We estimate the quantization sensitivity of each layer by calibrating the pre-trained model with the calibration images.
By feeding the calibration images to the pre-trained floating-point network, we collect the standard deviation of the activations, which can be used as a metric to estimate the layer-wise quantization sensitivity~\cite{hong2022cadyq}.
Using the calibrated sensitivity of each layer, we build a layer-to-bit mapping module with sensitivity thresholds that decreases the bit-width when the layer-wise sensitivity is lower than the lower bound, and vice versa.
\begin{equation}
    b^k_L = L2B(s^k) =
    \begin{cases}
    -1, & s^k < l_{l2b}, \\
    +1,  & s^k > u_{l2b}, \\
    0, & otherwise.
    \end{cases}
    \label{eq:l2b}
\end{equation} 
where $s^k$ is the sensitivity of $k$-th convolutional layer obtained by collecting average standard deviation of activations.
We use the $p_L$-th and $(100-p_L)$-th percentiles of the obtained sensitivity measures to initialize the sensitivity thresholds $l_{l2b}$ and $u_{l2b}$.
Since sensitivity measures can be precalculated before test time, the layer-wise bit factor can as well be determined before test time.
During inference, image-wise factor is adaptively obtained and then added to the pre-determined layer-wise factors.
To enable a further better performance-complexity trade-off, we set the layer-wise bit factors as learnable parameters, which are updated using calibration images for a small number of iterations.

\subsection{Bit-aware clipping}
For quantization clipping range, we first initialize the lower and upper bounds for each quantizer with the minimum and maximum statistics~\cite{Jacob_2018_CVPR} collected by running calibration images through the FP network, using the exponential moving average (EMA).
However, since bit-widths of quantizers are updated based on the input image and layer, the clipping range should also be updated.
When the bit-width is decreased, the number of quantization levels are reduced, thus the optimal clipping range changes.
Therefore, we adjust the clipping ranges with respect to the bit-width assigned to each layer.
Inspired by OMSE~\cite{choukroun2019low}, we adjust $l_a, u_a$ to minimize the L2 distance between the FP tensor and the quantized tensor of the given layer-wise bit-width $b$.
\begin{equation}
    \epsilon_* = \operatorname*{argmin}_\epsilon ||(\bm{X}-Q(\bm{X}; b, \epsilon\cdot l_a, \epsilon\cdot u_a)||_{2},
    \label{eq:bac}
\end{equation}
where $\epsilon_*$ is searched ranging from $1.0$ to $0.0$ for 100 steps and the final adjusted clipping range is $[\epsilon_*\cdot l_a, \epsilon_*\cdot u_a]$.
The process finds a better initial clipping range considering the layer-wise different bit-widths.

\begin{algorithm}[t]
\caption{Overall process of AdaBM} 
\label{alg}
\textbf{Input:} Pre-trained 32-bit SR network $\mathcal{P}$ of $K$ layers, calibration dataset $\mathcal{D}_{cal}=\{I_{LR}^i\}_{i=1}^N$. \\
\textbf{Output:} Adaptively quantized network $\mathcal{Q}$.   
\begin{algorithmic}
\State \textbf{Initialization Phase:}
\For{$i=1,\cdots,N$}
    \State Measure image complexity $c(I_{LR}^i)$
    \State Measure layer sensitivities $\{s^k\}^K_{k=1}$ from $\mathcal{P}(I_{LR}^i)$
\EndFor
\State Given $\{c(I_{LR}^i)\}^N_{i=1}$, initialize I2B thresholds $l_{i2b}, u_{i2b}$
\State Given $\{s^k\}^K_{k=1}$, obtain L2B thresholds $l_{l2b}, u_{l2b}$
\State Initialize layer-wise bit factors $\{b_L^k\}_{k=1}^K$ using \cref{eq:l2b}
\State Initialize clipping range thresholds $\{l_a^k, u_a^k\}_{k=1}^K$ based on layer-wise bits using \cref{eq:bac}
\State \textbf{Finetuning Phase:}
\For{epoch$=1, \cdots,\#$epochs}
    \State Update $\{l_{i2b}, u_{i2b}\}$ and $\{b_L^k\}^K_{k=1}$ using \cref{eq:loss-mapping}
    \State Update $\{l_a^k, u_a^k, u_w^k\}^K_{k=1}$ using \cref{eq:loss-range}
\EndFor
\end{algorithmic}
\end{algorithm}

\begin{table*}[t]
\renewcommand{\arraystretch}{1.1}
\centering
\aboverulesep=0ex
\belowrulesep=0ex
\resizebox{0.99\textwidth}{!}{
    \begin{tabular}{l|cc|c| c| cc cc cc}
        \toprule
        \multirow{2}{*}{Model} & \multirow{2}{*}{QAT} & \multirow{2}{*}{GT} & {Process} & \multirow{2}{*}{W / A} & 
        \multicolumn{2}{c}{Urban100} & \multicolumn{2}{c}{Test2K} & \multicolumn{2}{c}{Test4K}\\
        \cmidrule(lr){6-7} \cmidrule(lr){8-9} \cmidrule(lr){10-11}
        & & & Time & & FAB & PSNR / SSIM & FAB & PSNR / SSIM & FAB & PSNR / SSIM \\
        \midrule
        EDSR & - & - & - & 32 / 32 & 32.0 & 26.04 / 0.784 & 32.0 & 27.71 / 0.782 & 32.0 & 28.80 / 0.814\\
        \midrule
        EDSR-CADyQ & \cm & \cm & 40 hrs & 8 / 6MP & 6.6 & 25.98 / 0.784 & 6.1 & 27.69 / 0.782 & 6.0 & 28.79 / 0.814 \\
        EDSR-CABM  & \cm & \cm & 70 hrs & 8 / 6MP & 5.8 & 25.90 / 0.782 & 5.6 & 27.65 / 0.781 & 5.5 & 28.73 / 0.812 \\
        EDSR-AdaBM (\textbf{Ours}) & \xm & \xm & \textbf{71 sec} & 8 / 6MP & \textbf{5.7} & \textbf{25.96 / 0.782} & \textbf{5.3} & \textbf{27.65 / 0.779} & \textbf{5.2} & \textbf{28.71 / 0.809} \\
        \midrule
        EDSR-CADyQ & \cm & \cm & 40 hrs& 4 / 4MP & 4.9 & 25.12 / 0.753 & 4.9 & 27.43 / 0.771 & 4.9 & 28.49 / 0.803 \\
        EDSR-CABM & \cm & \cm & 70 hrs& 4 / 4MP & 4.4 & 24.98 / 0.746 & 4.4 & 27.33 / 0.767 & 4.4 & 28.36 / 0.798 \\
        EDSR-AdaBM (\textbf{Ours}) & \xm & \xm & \textbf{71 sec}& 4 / 4MP & \textbf{4.3} & \textbf{25.49 / 0.759} & \textbf{3.9} & \textbf{27.40 / 0.758} & \textbf{3.8} & \textbf{28.39 / 0.784}\\
        \midrule
        SRResNet & - & - & - & 32 / 32 & 32.0 & 25.86 / 0.779 & 32.0 & 27.64 / 0.781 & 32.0 & 28.72 / 0.813 \\
        \midrule
        SRResNet-CADyQ & \cm & \cm & 53 hrs & 8 / 6MP & 6.4 & 25.89 / 0.779 & 6.2 & 27.65 / 0.780 & 6.2 & 28.73 / 0.812\\
        SRResNet-CABM  & \cm & \cm & 93 hrs & 8 / 6MP & 5.8 & 25.79 / 0.776 & 5.6 & 27.60 / 0.779 & 5.6 & 28.67 / 0.811 \\
        SRResNet-AdaBM (\textbf{Ours}) & \xm & \xm & \textbf{92 sec} & 8 / 6MP & \textbf{5.6} & \textbf{25.72 / 0.773} & \textbf{5.2} & \textbf{27.55 / 0.777} & \textbf{5.1} & \textbf{28.62 / 0.809} \\
        \midrule
        SRResNet-CADyQ & \cm & \cm & 53 hrs & 4 / 4MP & 4.1 & 25.39 / 0.761 & 4.1 & 27.38 / 0.771 & 4.1 & 28.46 / 0.804 \\
        SRResNet-CABM & \cm & \cm & 93 hrs & 4 / 4MP & 3.8 & 25.42 / 0.764 & 3.8 & 27.46 / 0.774 & 3.6 & 28.52 / 0.806 \\
        SRResNet-AdaBM (\textbf{Ours}) & \xm & \xm & \textbf{92 sec} & 4 / 4MP & \textbf{4.2} & \textbf{25.32 / 0.757} & \textbf{3.9} & \textbf{27.31 / 0.766} & \textbf{3.9} & \textbf{28.25 / 0.782} \\
        \bottomrule
    \end{tabular}
}
\vspace{-0.2cm}
\caption{ \textbf{Comparisons with adaptive quantization methods on SR} on EDSR and SRResNet of scale 4. 
Previous methods require extensive quantization-aware training (QAT) process with ground-truth (GT) HR images, which result in long process time.
}
\vspace{-4mm}
\label{tab:exp-qat}
\end{table*}

\subsection{Finetuning}
Calibrating the pre-trained model with the calibration images derives a decent initial point for the quantized network. 
Nevertheless, we fine-tune the mapping with the calibration images to obtain a better bit mapping for input image and layers.
As layer-wise bit factors are pre-determined and fixed during test time, we directly fine-tune the layer-wise bit factors $\{b_L^k\}_{k=1}^K$.
In contrast, image-wise bit factors are adapted at test time, thus we fine-tune the mapping module parameterized by threshold values $l_{i2b}, u_{i2b}$.
However, the thresholding function of the image-to-bit mapping is not differentiable.
Accordingly, we approximate the thresholding function with tanh~\cite{gong2019differentiable} during back-propagation as:
\begin{equation}
    \frac{\partial {b}_I^j}{\partial u_{i2b}} \approx \frac{\partial \tilde{{b}}_I^j}{\partial u_{i2b}} \;\; \text{and} \;\;
    \frac{\partial {b}_I^j}{\partial l_{i2b}} \approx \frac{\partial \tilde{{b}}_I^j}{\partial l_{i2b}},
    \label{eq:STE-bit}
\end{equation}
\begin{equation}
    \text{where}\;\; \tilde{{b}}_I^j = tanh({c}(I_{LR}^j)-\frac{u_{i2b}+l_{i2b}}{2}).
    \label{eq:STE-bit2}
\end{equation}
Also, we update the clipping ranges for quantizing activations ($l_a, u_a$) and weights ($-u_w, u_w$). 
Since the quantizer includes the rounding function that is not differentiable, we employ STE~\cite{bengio2013estimating} to approximate the rounding function as identity function during back-propagation as: 
\begin{equation}
    \frac{\partial \bm{X}_q}{\partial l_a} \approx 
    \begin{cases}
    1, & \bm{X} < l_a \\
    0, & \bm{X} \geq l_a \\
    \end{cases}
    ,
    \frac{\partial \bm{X}_q}{\partial u_a} \approx
    \begin{cases}
    1, & \bm{X} > u_a \\
    0, & \bm{X} \leq u_a \\
    \end{cases}
    ,
    \label{eq:STE-activation}  
\end{equation}
\begin{equation}
     \frac{\partial \bm{W}_q}{\partial u_w} \approx
    \begin{cases}
    1, & |\bm{W}| > u_w \\
    0, & |\bm{W}| \leq u_w \\
    \end{cases}
    .
    \label{eq:STE-weight}    
\end{equation}
The network weights of the SR network are frozen and only the quantization parameters are optimized using reconstruction losses from supervision of the pre-trained FP network $\mathcal{P}$ following~\cite{tu2023toward}.
Reconstruction losses are given as:
\begin{equation}
\begin{split}
    \mathcal{L}_{pix}=\frac{1}{N} \sum_i^N ||\mathcal{P}(I_{LR}^i)- \mathcal{Q}(I_{LR}^i)||_1, \\
    \mathcal{L}_{skt}=\frac{1}{N\cdot K}\sum_i^N \sum_k^K ||\frac{F_{\mathcal{P}}^{i,k}}{||F_{\mathcal{P}}^{i,k}||_2}- \frac{F_{\mathcal{Q}}^{i,k}}{||F_{\mathcal{Q}}^{i,k}||_2} ||_2,
    \label{eq:loss-rec}
\end{split}
\end{equation}
where $F_{\mathcal{P}}^{i,k}$ and $F_{\mathcal{Q}}^{i,k}$ are the outputs of $k$-th layer of $\mathcal{P}$ and $\mathcal{Q}$ and $N$ denotes the batch size. 
Such supervision only requires LR inputs and no ground-truth HR images, allowing us to fine-tune quantization parameters only with LR calibration images.
Additionally, since we do not want overly high bit-widths to be assigned to increase the reconstruction accuracy, we regularize the bit-width to balance computational cost and accuracy.
\begin{equation}
    \mathcal{L}_{bit} = max(\frac{1}{N\cdot K} \sum_{i}^N \sum_{k}^K b^{i,k}-b_{tar}, 0),
    \label{eq:loss-bit}
\end{equation}
where 
we set target bit-width $b_{tar}$ as the static bit-width $b_{base}$.
We use bit regularization together with the reconstruction loss to update the bit mapping parameters:
\begin{equation}
    \mathcal{L}_{pix} + \lambda_{skt}\mathcal{L}_{skt} + \lambda_{bit}\mathcal{L}_{bit}.
    \label{eq:loss-mapping}    
\end{equation}
For clipping ranges, we only apply the reconstruction loss:
\begin{equation}
    \mathcal{L}_{pix} + \lambda_{skt}\mathcal{L}_{skt},
    \label{eq:loss-range} 
\end{equation}
Moreover, instead of optimizing all quantization parameters at once,
we iteratively update the bit mapping parameters, clipping ranges for weights, then clipping ranges for activations.
While the bit mapping parameters are updated, the clipping ranges of weights and activations are frozen.
Afterwards, we freeze the bit mappings, 
then update the clipping ranges, such that clipping ranges are optimized for the current bit assignment policy.
With such an update scheme, we achieve saturation within a small number of iterations.
The overall process is summarized in \Cref{alg}.
\section{Experiments}
\label{sec:experiments}

\subsection{Implementation details}
First, we build the calibration dataset by randomly sampling 100 LR images from the DIV2K~\cite{agustsson2017ntire} training dataset.
The calibration images are used to calibrate and fine-tune our bit mappings and the quantization ranges.
The calibration is done for one epoch with a batch size of 16.
The quantization range for activations is initialized using MinMax~\cite{Jacob_2018_CVPR} and weights OMSE~\cite{choukroun2019low}.
The hyperparameters for calibrating the bit mapping modules, $p_I$ and $p_L$ are set to 10 and 30. 
Then, after freezing the network weights, we fine-tune only the clipping ranges for the weights and activations and the parameters for mapping modules for 10 epochs with a batch size of 2 using the Adam optimizer~\cite{kingma2014adam}.
The initial learning rate for the activation clipping ranges, the weight clipping ranges, the layer-wise bit factors, and the image-to-bit mapping module parameters are set as 0.01, 0.01, 0.01, and 0.1, respectively.
Each learning rate is decayed by 0.9 every epoch.
We balance the different loss terms with a loss weight of $\lambda_{skt}=10, \lambda_{bit}=50$.
For a fair comparison with existing adaptive quantization methods, we follow the common practice~\cite{hong2022cadyq,tian2023cabm,zhong2022ddtb,Li2020pams} of quantizing weights and activations of the body part of the SR model.
Also, we follow existing adaptive methods~\cite{hong2022cadyq,tian2023cabm} to apply adaptive quantization to activations and static quantization to weights.

\begin{table*}[t]
\renewcommand{\arraystretch}{1.1}
\centering
\aboverulesep=0ex
\belowrulesep=0ex
\resizebox{0.99\textwidth}{!}{
    \begin{tabular}{l|c|c| cc cc cc cc}
        \toprule
        \multirow{2}{*}{Model} & \multirow{2}{*}{FT} & \multirow{2}{*}{W / A} & 
        \multicolumn{2}{c}{Set5} & \multicolumn{2}{c}{Set14} & \multicolumn{2}{c}{BSD100} & \multicolumn{2}{c}{Urban100} \\
        \cmidrule(lr){4-5} \cmidrule(lr){6-7} \cmidrule(lr){8-9} \cmidrule(lr){10-11}& 
         && FAB & PSNR / SSIM & FAB & PSNR / SSIM & FAB & PSNR / SSIM & FAB & PSNR / SSIM \\
        \midrule
        EDSR &-&32 / 32 & 32.0 & 32.10 / 0.894 & 32.0 & 28.58 / 0.781 & 32.0 & 27.56 / 0.736 & 32.0 & 26.04 / 0.785\\
        \midrule
        EDSR-MinMax        &\xm& 6 / 6 & 6.0 & 31.56 / 0.866 & 6.0 & 28.26 / 0.760 & 6.0 & 27.29 / 0.714 & 6.0 & 25.76 / 0.760\\
        EDSR-Percentile    &\xm& 6 / 6 & 6.0 & 24.30 / 0.793 & 6.0 & 24.31 / 0.728 & 6.0 & 24.68 / 0.700 & 6.0 & 21.93 / 0.696\\
        EDSR-MinMax+FT     &\cm& 6 / 6 & 6.0 & 31.61 / 0.870 & 6.0 & 28.31 / 0.762 & 6.0 & 27.34 / 0.718 & 6.0 & 25.81 / 0.763\\
        EDSR-Percentile+FT &\cm& 6 / 6 & 6.0 & 27.23 / 0.832 & 6.0 & 25.89 / 0.747 & 6.0 & 25.82 / 0.716 & 6.0 & 23.35 / 0.723\\
        EDSR-PTQ4SR        &\cm& 6 / 6 & 6.0 & 31.80 / 0.884 & 6.0 & 28.34 / 0.768 & 6.0 & 27.37 / 0.722 & 6.0 & 25.79 / 0.769\\
        EDSR-AdaBM (\textbf{Ours}) &\cm& 6 / 6MP & \textbf{5.7} & \textbf{31.92} / \textbf{0.887} & \textbf{5.6} & \textbf{28.47} / \textbf{0.777} & \textbf{5.4} & \textbf{27.47} / \textbf{0.731} & \textbf{5.7} & \textbf{25.89} / \textbf{0.778}\\
        \midrule
        EDSR-MinMax        &\xm& 4 / 4 & 4.0 & 26.83 / 0.624 & 4.0 & 25.04 / 0.546 & 4.0 & 24.57 / 0.503 & 4.0 & 23.12 / 0.536\\
        EDSR-Percentile    &\xm& 4 / 4 & 4.0 & 24.03 / 0.776 & 4.0 & 23.95 / 0.712 & 4.0 & 24.42 / 0.687 & 4.0 & 21.62 / 0.677\\
        EDSR-MinMax+FT     &\cm& 4 / 4 & 4.0 & 28.97 / 0.821 & 4.0 & 26.47 / 0.721 & 4.0 & 26.24 / 0.687 & 4.0 & 23.46 / 0.674\\
        EDSR-Percentile+FT &\cm& 4 / 4 & 4.0 & 27.01 / 0.819 & 4.0 & 25.71 / 0.736 & 4.0 & 25.69 / 0.707 & 4.0 & 23.18 / 0.707\\
        EDSR-PTQ4SR        &\cm& 4 / 4 & 4.0 & 30.51 / 0.836 & 4.0 & 27.62 / 0.735 & 4.0 & 26.88 / 0.693 & 4.0 & 24.92 / 0.721\\
        EDSR-AdaBM (\textbf{Ours}) &\cm& 4 / 4MP & \textbf{3.8} & \textbf{31.02} / \textbf{0.860} & \textbf{3.7} & \textbf{27.87} / \textbf{0.751} & \textbf{3.5} & \textbf{26.91} / \textbf{0.700} & \textbf{3.7} & \textbf{25.11} / \textbf{0.736} \\
        \midrule
        RDN &-&32 / 32 & 32.0 & 32.24 / 0.895 & 32.0 & 28.67 / 0.784 & 32.0 & 27.63 / 0.739 & 32.0 & 26.29 / 0.793\\
        \midrule
        RDN-MinMax        &\xm& 6 / 6 & 6.0 & 30.59 / 0.863 & 6.0 & 27.54 / 0.752 & 6.0 & 26.65 / 0.703 & 6.0 & 24.79 / 0.733\\
        RDN-Percentile    &\xm& 6 / 6 & 6.0 & 18.87 / 0.778 & 6.0 & 18.33 / 0.667 & 6.0 & 19.88 / 0.651 & 6.0 & 16.81 / 0.632\\
        RDN-MinMax+FT     &\cm& 6 / 6 & 6.0 & 31.16 / 0.873 & 6.0 & 27.92 / 0.762 & 6.0 & 27.03 / 0.716 & 6.0 & 25.23 / 0.749\\
        RDN-Percentile+FT &\cm& 6 / 6 & 6.0 & 21.32 / 0.812 & 6.0 & 20.74 / 0.702 & 6.0 & 21.87 / 0.677 & 6.0 & 18.67 / 0.670\\
        RDN-PTQ4SR        &\cm& 6 / 6 & 6.0 & 30.73 / 0.877 & 6.0 & 27.60 / 0.765 & 6.0 & 26.85 / 0.720 & 6.0 & 25.08 / 0.756\\
        RDN-AdaBM (\textbf{Ours}) &\cm& 6 / 6MP & \textbf{5.7} & \textbf{31.56} / \textbf{0.881} & \textbf{5.6} & \textbf{28.14} / \textbf{0.769} & \textbf{5.5} & \textbf{27.20} / \textbf{0.722} & \textbf{5.7} & \textbf{25.31} / \textbf{0.755}\\
        \midrule
        RDN-MinMax        &\xm& 4 / 4 & 4.0 & 25.91 / 0.632 & 4.0 & 24.22 / 0.549 & 4.0 & 24.29 / 0.530 & 4.0 & 22.24 / 0.523 \\
        RDN-Percentile    &\xm& 4 / 4 & 4.0 & 18.83 / 0.771 & 4.0 & 18.28 / 0.662 & 4.0 & 19.83 / 0.646 & 4.0 & 16.77 / 0.625 \\
        RDN-MinMax+FT     &\cm& 4 / 4 & 4.0 & 28.50 / 0.810 & 4.0 & 26.15 / 0.703 & 4.0 & 26.00 / 0.673 & 4.0 & 23.35 / 0.673 \\
        RDN-Percentile+FT &\cm& 4 / 4 & 4.0 & 21.24 / 0.798 & 4.0 & 20.68 / 0.690 & 4.0 & 21.85 / 0.668 & 4.0 & 18.64 / 0.652 \\
        RDN-PTQ4SR        &\cm& 4 / 4 & 4.0 & 28.32 / 0.813 & 4.0 & 26.11 / 0.709 & 4.0 & 25.82 / 0.671 & 4.0 & 23.31 / 0.668 \\
        RDN-AdaBM (\textbf{Ours}) &\cm& 4 / 4MP & \textbf{3.8} & \textbf{28.71} / \textbf{0.808} & \textbf{3.7} & \textbf{26.30} / \textbf{0.707} & \textbf{3.6} & \textbf{26.10} / \textbf{0.672} & \textbf{3.8} & \textbf{23.38} / \textbf{0.663}\\
        \bottomrule
    \end{tabular}
}
\vspace{-0.2cm}
\caption{ \textbf{Comparisons with static quantization without QAT on SR.} We evaluate on EDSR of scale 4 that consists of 16 residual blocks (64 channels) and RDN of scale 4. 
}
\vspace{-4mm}
\label{tab:exp-ptq}
\end{table*}

\subsection{Comparison with adaptive quantization}
To demonstrate the effectiveness of our on-the-fly scheme for adaptive quantization, we compare it with existing adaptive quantization methods, CADyQ~\cite{hong2022cadyq} and CABM~\cite{tian2023cabm}, which we reproduce using the official codebase.
The methods are evaluated on representative SR networks: EDSR~\cite{lim2017enhanced} and SRResNet~\cite{ledig2017photo} of scale 4.
The methods are tested on large input datasets, Urban100~\cite{huang2015single}, Test2K, and Test4k~\cite{kong2021classsr}, in which Test2K and Test4K are produced by downsampling DIV8K~\cite{gu2019div8k} dataset.
Following existing methods~\cite{hong2022cadyq,tian2023cabm}, a large input image is split into small patches of size $96\times96$.
For evaluation metrics, we measure reconstruction accuracy using the peak signal-to-noise ratio (PSNR) and the structural similarity index (SSIM).
To compare the computational complexity of the quantized network, we report the feature average bit-width (FAB) that is averaged throughout the images of the test dataset.

As shown in \Cref{tab:exp-qat}, although our method does not employ a quantization-aware training (QAT) process using the full training pairs with GT, we achieve a performance comparable to the existing methods.
For example, our method achieves similar reconstruction accuracy with lower FAB on most of the settings.
The results demonstrate that our method can accelerate the processing time in hours to the processing level in seconds for the first time without sacrificing the performance.

\subsection{Comparison with static quantization}
We propose the first adaptive quantization method that does not require an extensive quantization-aware training (QAT) process.
To further validate the effectiveness of our adaptive approach, we compare it with existing static quantization methods that do not involve the QAT process.
We compare with commonly used PTQ methods MinMax~\cite{Jacob_2018_CVPR} and Percentile~\cite{li2019fully}, which calibrates the clipping parameters by using percentiles of weight/activation statistics (1\% for lower bound and 99\% for upper bound).
However, these methods suffer severe accuracy degradation when applied directly to SR networks.
Thus, we calibrate the quantized network with these methods and then fine-tune the quantization parameters using calibration images, denoted as MinMax+FT and Percentile+FT.
Furthermore, we compare with a more recent approach on SR that finetunes the quantization parameters: PTQ4SR~\cite{tu2023toward}, which we reproduce for comparison.
For a fair comparison with PTQ4SR, we follow their setting to apply static 8-bit quantization on head and tail parts, such that weights and activations of all convolutional layers are quantized.
We note that unlike previous methods that employ asymmetric quantization functions for both weights and activations, we apply symmetric quantization to the weights and asymmetric quantization to the activations.
We evaluate these methods on representative SR networks, EDSR~\cite{lim2017enhanced} that consists of 16 residual blocks with 64 channel dimensions, and a more extensive SR network, RDN~\cite{zhang2018residual} of scale 4.
As reported in \Cref{tab:exp-ptq}, our method, AdaBM, outperforms existing methods for both 6-bit and 4-bit settings with lower FAB.
This indicates that our method achieves a better trade-off between reconstruction accuracy and computational costs.
Further experiments to demonstrate the applicability of our method are provided in the supplementary document.

\subsection{Qualitative results}
\begin{figure*}[t]
\centering
\begin{center}
    \setlength{\tabcolsep}{0.07cm}
    \newcommand{\w}{0.15\linewidth}
    \begin{tabular}{ccccc}
        \centering
        \includegraphics[width=\w]{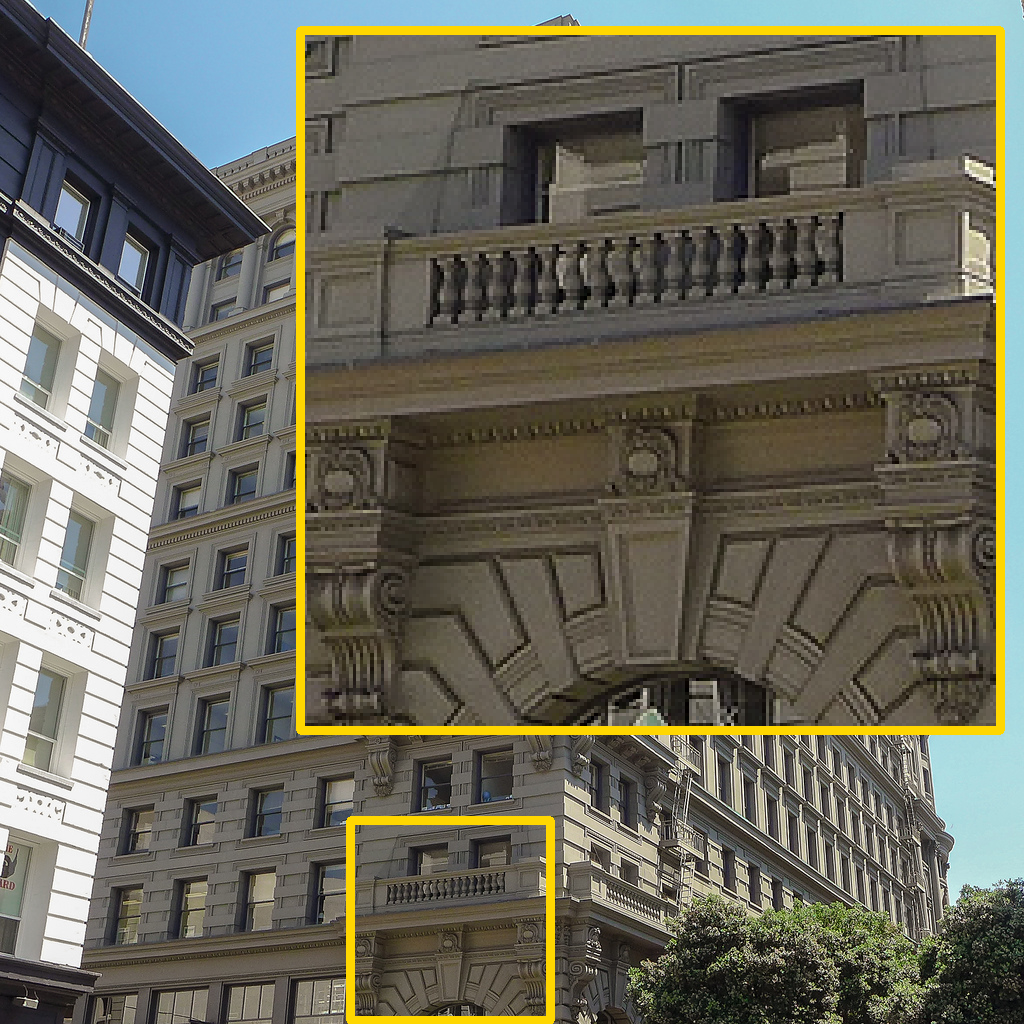} & \includegraphics[width=\w]{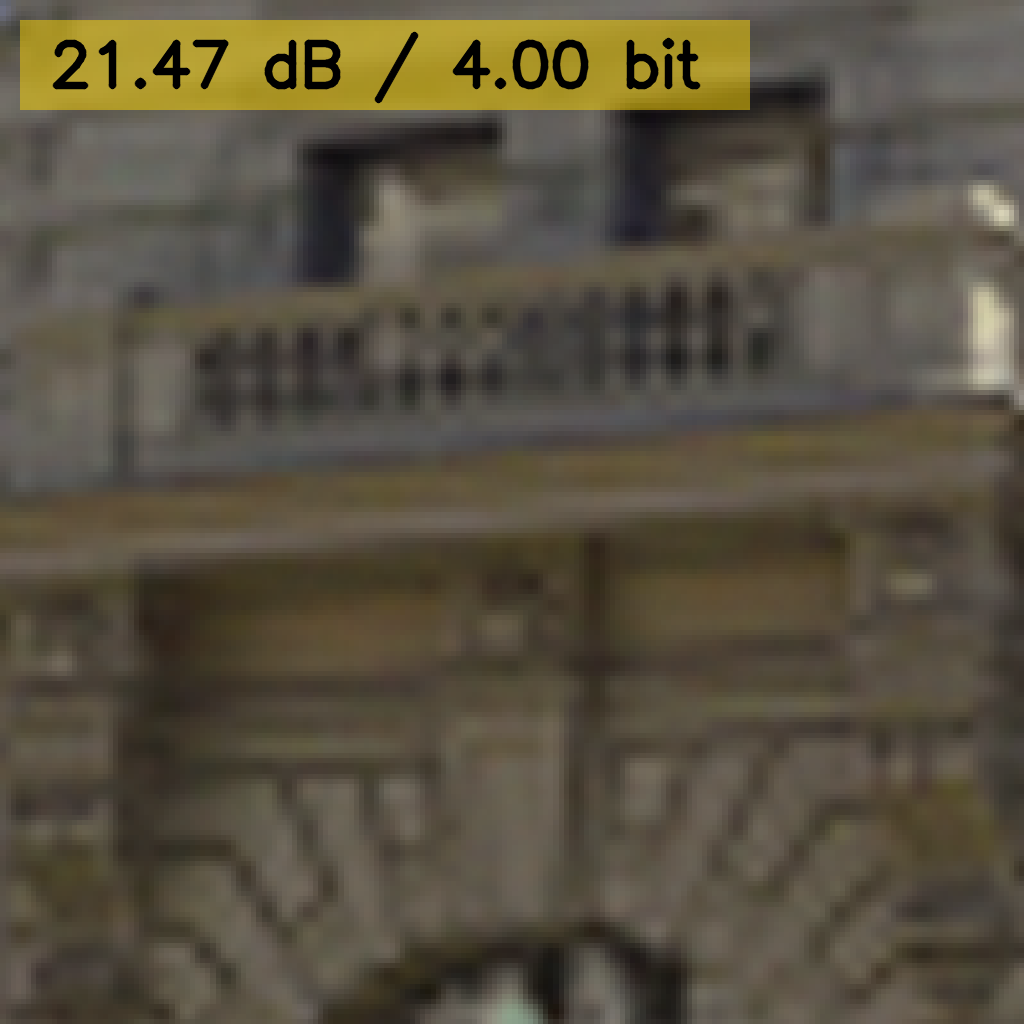} & \includegraphics[width=\w]{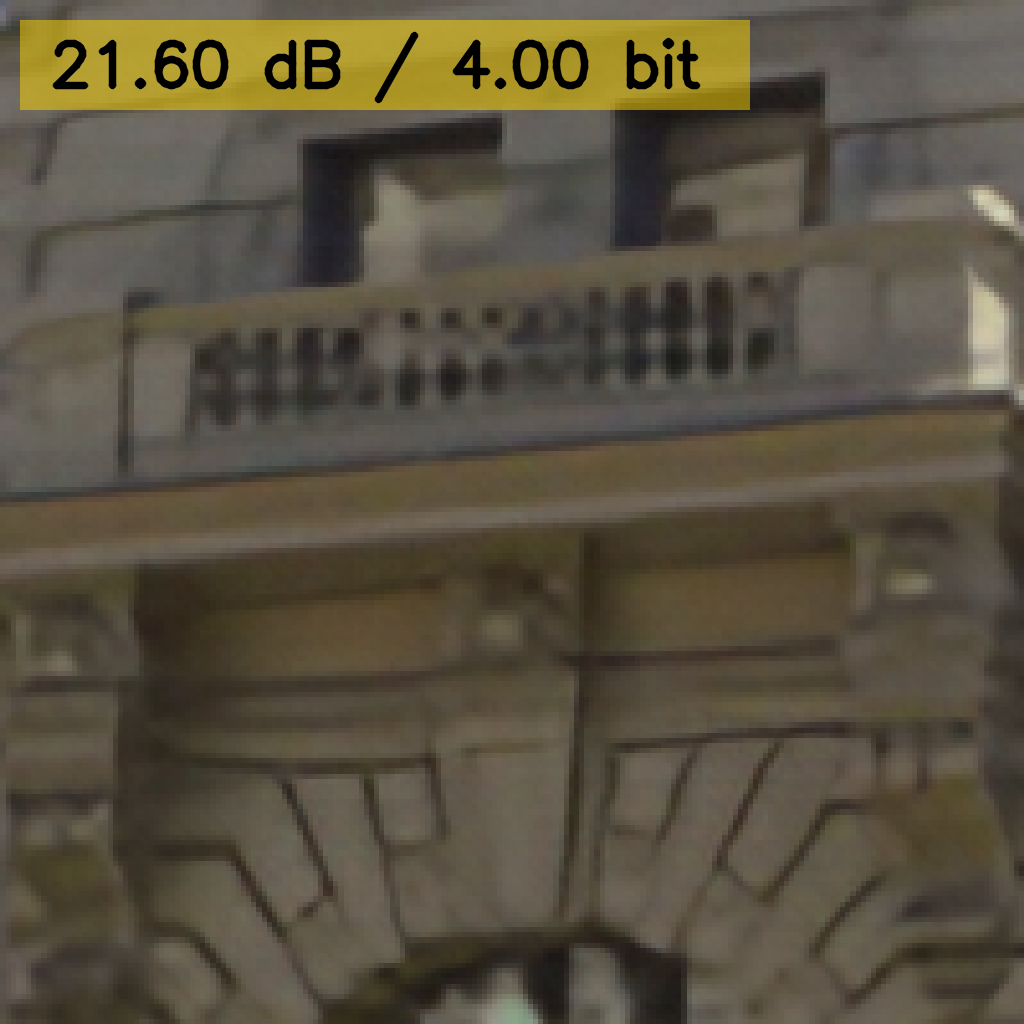} & \includegraphics[width=\w]{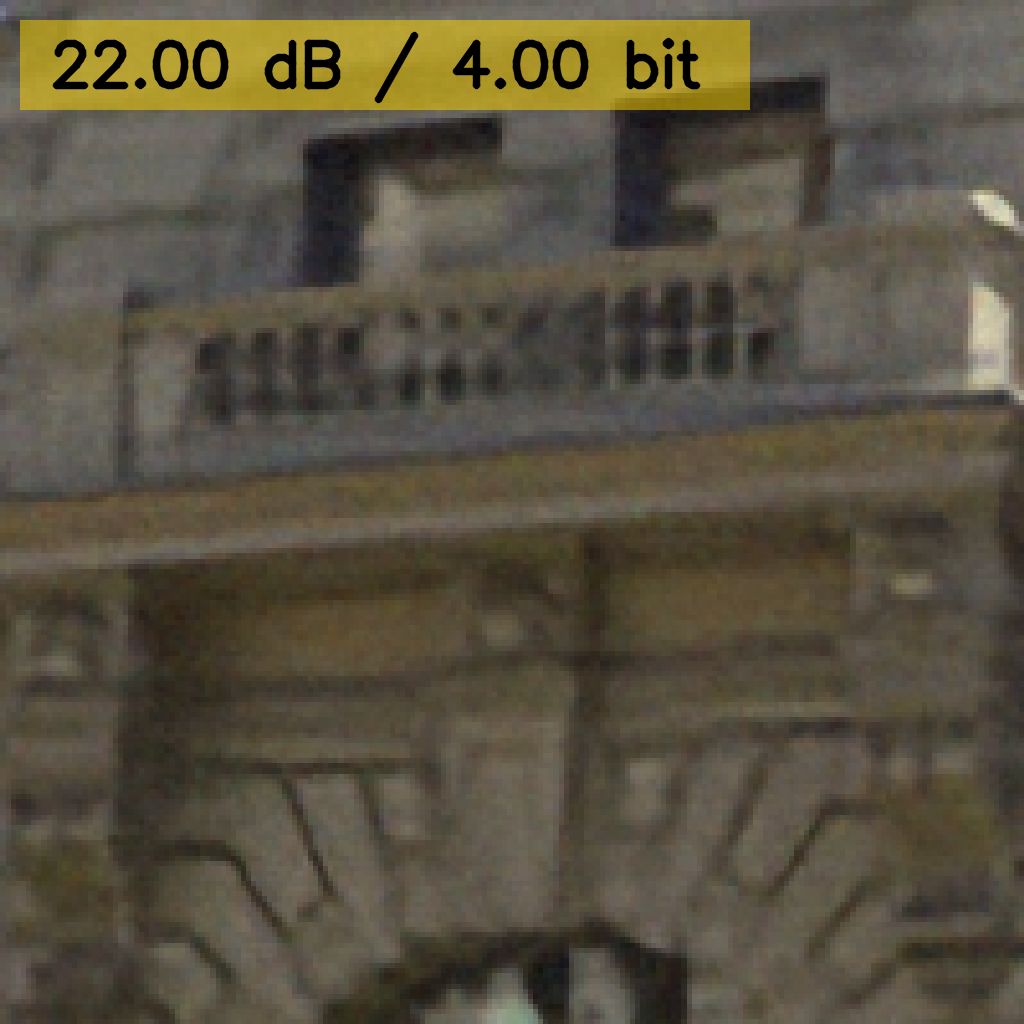} & \includegraphics[width=\w]{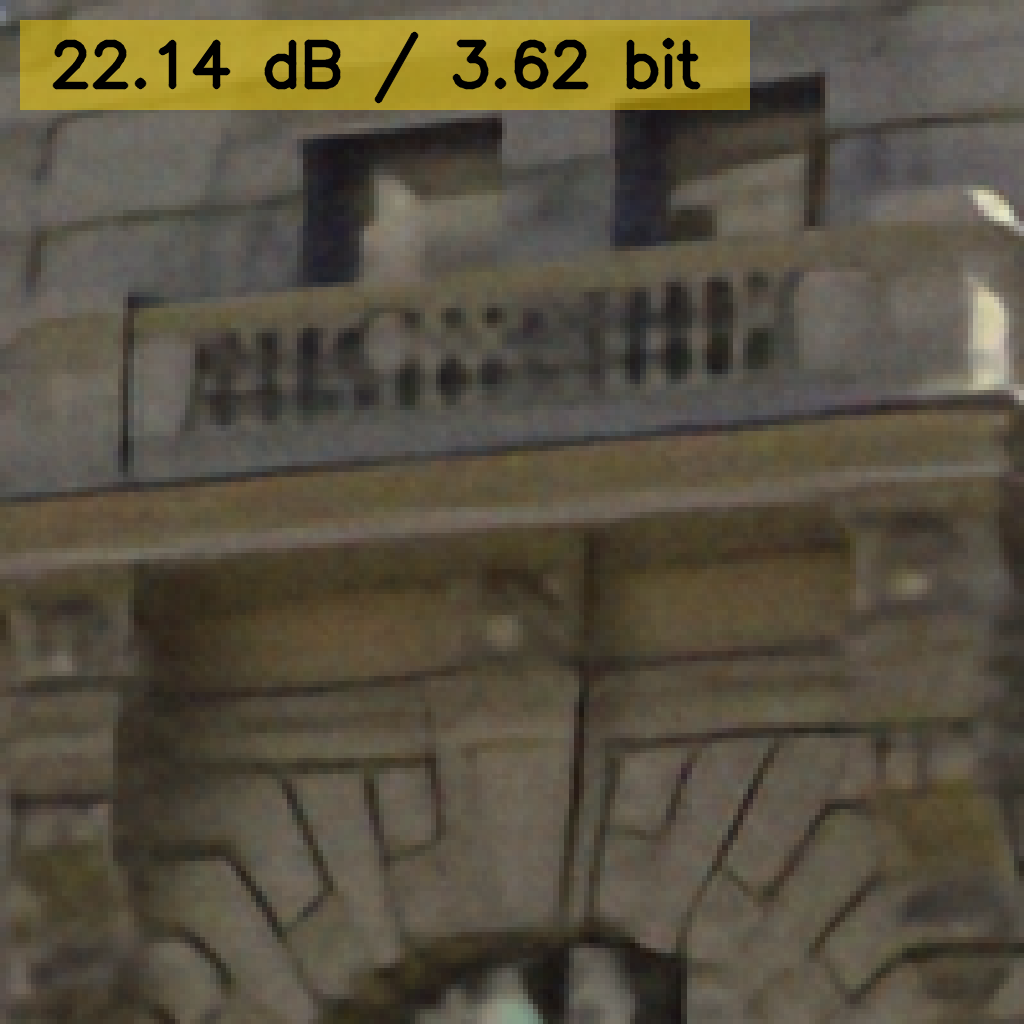} \\
        \small GT (img014) & \small MinMax+FT & \small Percentile+FT &\small PTQ4SR &\small AdaBM (Ours) \\
    \end{tabular}
    \vspace{-0.2cm}
    \caption{
    \textbf{Qualitative results} on Urban100 with 4-bit EDSR-based models.
    More results are provided in the supplementary document.
    }
    \vspace{-0.3cm}
    \label{fig:exp-qual}
\end{center}
\end{figure*}

\begin{figure*}
    \centering
    \begin{minipage}[b]{0.2\linewidth}
        \centering
        \includegraphics[width=\textwidth,height=86pt]{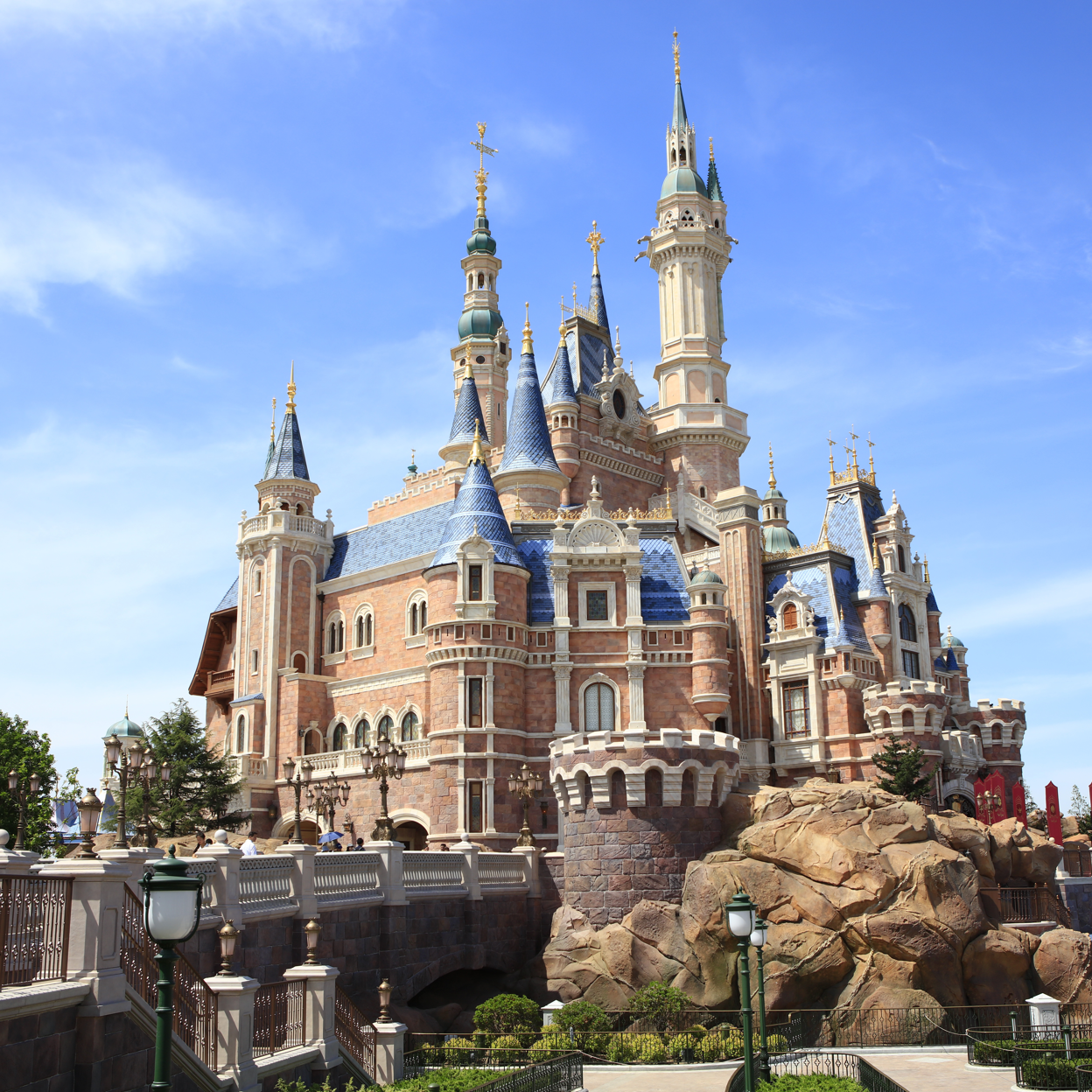}
    \end{minipage}
    \hspace{1mm}
    \begin{minipage}[b]{0.2\linewidth}
        \centering
        \includegraphics[width=\textwidth,height=86pt]{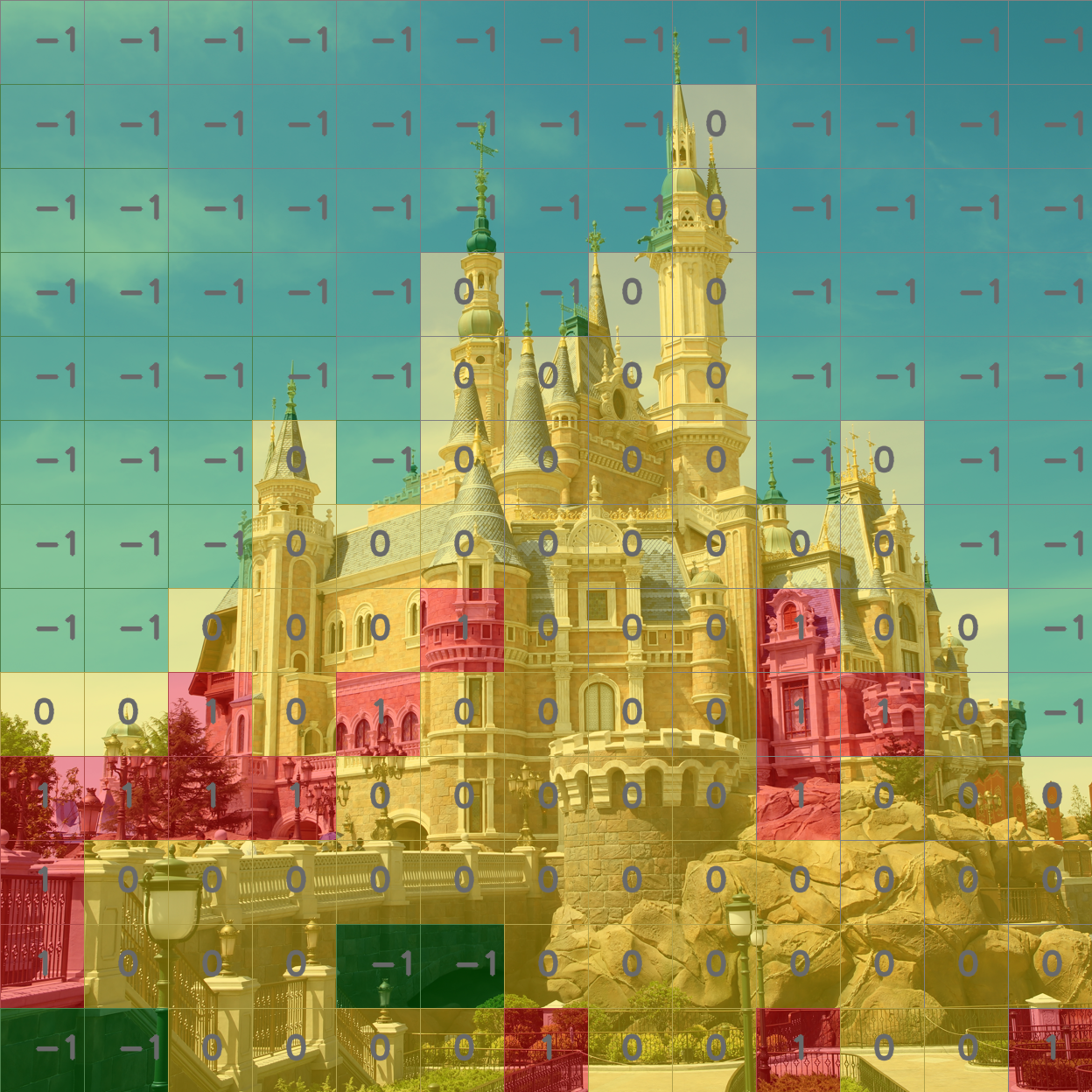}
    \end{minipage}
    \hspace{5mm}
    \begin{minipage}[b]{0.3\linewidth}
        \centering
        \includegraphics[height=37pt]{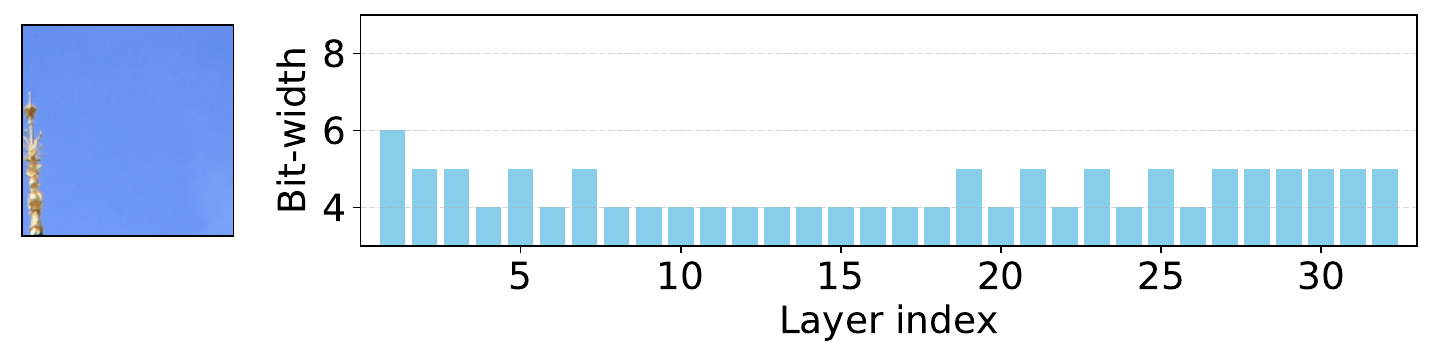}\par\vspace{2pt}
        \includegraphics[height=37pt]{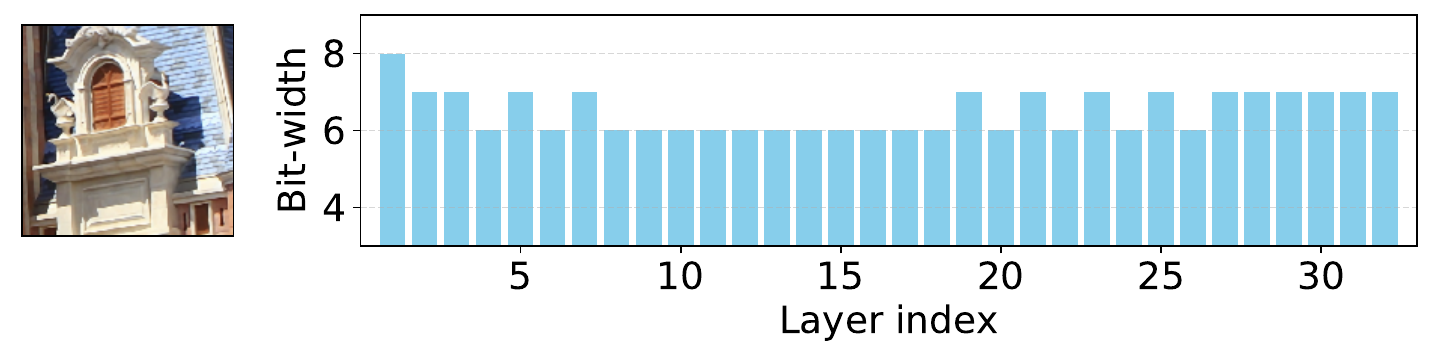}
    \end{minipage}
    \vspace{-0.2cm}
    \caption{
        \textbf{Visualization of adaptive bit-mapping of AdaBM.} 
        Results are from EDSR-AdaBM {(w8 a6MP)}.
    }
    \vspace{-3mm}
\label{fig:exp-vis}
\end{figure*}

We also evaluate our framework qualitatively and compare it with the existing quantization methods without QAT.
As shown in \Cref{fig:exp-qual}, our method produces visually better-reconstructed images with lower average bit-width. 
Compared to other methods, the outputs of our method retain more details.
Additionally, the adaptive allocation of our framework is visualized in \Cref{fig:exp-vis}.
Different bit-widths are assigned to different images and layers.
Image-wise allocation results show that higher bits are assigned to complex images with more structural information and lower bits for images of fewer details (\eg, sky).

\subsection{Ablation study}
We conduct an ablation study on each attribute of our framework to investigate the effect of the image-wise bit-mapping, layer-wise bit-mapping, bit-aware clipping, and fine-tuning with calibration images.
As presented in \Cref{tab:exp-ablation}, using both image and layer-wise bit mapping is essential to achieve higher reconstruction accuracy with lower computational costs (FAB).
Furthermore, bit-aware clipping gives an additional gain in reconstruction accuracy.
Also, finetuning with the calibration images allows for a better accuracy-complexity Pareto frontier.

\subsection{Complexity analysis}
We analyze the computational complexities and processing time of our framework compared to existing methods in \Cref{tab:exp-complexity}.
In terms of processing time, ours is over $\times2000$ faster than existing adaptive quantization methods.
Also, the processing time is even shorter than that of several static quantization methods without QAT, showing our adaptive bit-mapping scheme has little or no processing time overhead over static quantization.
Moreover, we measure the computational complexity: the model size required to store the model weights and the average bitOPs for Urban100 images.
The results show that AdaBM consumes less bitOPs and model storage size compared to existing methods.

\begin{table}[t]
    \renewcommand{\arraystretch}{1.2}
    \centering
    \aboverulesep=0ex
    \belowrulesep=0ex
    \resizebox{0.47\textwidth}{!}{
    \begin{tabular}{cccc|cc|cc}
        \toprule
        I2B & L2B & BaC & FT & FAB & PSNR/SSIM & FAB & PSNR/SSIM \\
        \midrule
        - & - & - & - &        4.0 & 28.70 / 0.734  & 4.0 & 24.27 / 0.628\\
        \cm & \cm & \cm & - &  4.2 & 30.00 / 0.825 & 4.2 & 24.78 / 0.705\\
        - & - & - & \cm &      4.0 & 29.02 / 0.823 & 4.0 & 23.51 / 0.677\\
        \cm & - & - & \cm&     4.2 & 29.18 / 0.804 & 4.1 & 23.80 / 0.662\\
        - & \cm & - & \cm&     4.0 & 30.47 / 0.836 & 4.0 & 24.89 / 0.714\\
        - & - & \cm & \cm&     4.0 & 30.68 / 0.837 & 4.0 & 25.12 / 0.724\\
        \cm & \cm & \cm & \cm& \textbf{3.8} & \textbf{31.02} / \textbf{0.860} & \textbf{3.7} & \textbf{25.11} / \textbf{0.736}\\
        \bottomrule
    \end{tabular}
    }
    \vspace{-0.2cm}
    \caption{ \textbf{Ablation study on each attribute of AdaBM} evaluated on Set5/Urban100 with 4-bit EDSR ($\times$4). 
    I2B and L2B respectively denote the image-wise and layer-wise bit-mapping module. BaC refers to the bit-aware clipping and FT refers to finetuning.
    }
    \label{tab:exp-ablation}
\end{table}

\begin{table}[t]
    \centering
    \renewcommand{\arraystretch}{1.2}
    \aboverulesep=0ex
    \belowrulesep=0ex
    \resizebox{0.47\textwidth}{!}{
        \begin{tabular}{l|c|c|ccr}
            \toprule 
            \multirow{2}{*}{Method} & \multirow{2}{*}{Adaptive} & \multirow{2}{*}{FQ}& Process & Model Size  & \multirow{2}{*}{BitOPs} \\
            & & & Time & ($r_{comp}$)\\
            \midrule 
            EDSR & - & - & - & 1517.6K (0.0\%) & 527.0T\\
            \midrule
            CADyQ         &\cm&\xm& 40 hrs & 489.2K (67.8\%) & 82.6T\\
            CABM          &\cm&\xm& 70 hrs & 485.5K (68.0\%) & 82.4T\\
            AdaBM (\textbf{Ours}) &\cm&\xm&  71 sec & 485.5K (68.0\%) & 81.6T\\
            \midrule
            MinMax+FT     &\xm&\cm&  75 sec & 305.5K (79.9\%) & 9.2T\\
            Percentile+FT &\xm&\cm& 101 sec & 305.5K (79.9\%) & 9.2T\\
            PTQ4SR        &\xm&\cm& 124 sec & 305.5K (79.9\%) & 9.2T\\
            AdaBM (\textbf{Ours}) &\cm&\cm&  76 sec & 305.5K (79.9\%) & 9.0T\\
            \bottomrule
        \end{tabular}
    }
    \vspace{-0.2cm}
    \caption{ \textbf{Complexity analysis for different quantization methods on SR.} 
    Metrics are reported for quantizing EDSR ($\times$4). 
    FQ denotes whether the network is fully quantized.
    The process time is measured on a single 2080Ti GPU.
    }
    \vspace{-5mm}
    \label{tab:exp-complexity}
\end{table}

\section{Conclusion}
\label{sec:conclusion}
In this paper, we present the first adaptive bit-mapping pipeline for image super-resolution that learns the bit-mapping policies on the fly.
Our bit allocation is formulated by two policies of adapting the bit-width to a higher bit for sensitive layers and complex input images.
Layers and images are directly mapped to bit-widths, in which the mapping modules are calibrated and fine-tuned using calibration images (\ie, a small subset of LR images without corresponding HR images).
The results demonstrate that our method achieves on-par accuracy with previous methods while the processing time is significantly reduced.

\paragraph{Acknowledgment}
This work was supported in part by the IITP grants [No. 2021-0-01343, Artificial Intelligence Graduate School Program (Seoul National University), No.2021-0-02068, and No.2023-0-00156], the NRF grant [No.2021M3A9E4080782] funded by the Korean government (MSIT).

{
    \small
    \bibliographystyle{ieeenat_fullname}
    \bibliography{main}
}

\onecolumn
\clearpage
\setcounter{section}{0}
\renewcommand{\thesection}{\Alph{section}}

\setcounter{table}{0}
\renewcommand{\thetable}{S\arabic{table}}
\setcounter{figure}{0}
\renewcommand{\thefigure}{S\arabic{figure}}

{
    \centering
    \Large{\textbf{\thetitle}}\\
    \vspace{0.1cm}\Large{Supplementary Material} \\
    \vspace{0.2cm}
    \large{Cheeun Hong$^1$ \hspace{3cm} Kyoung Mu Lee$^{1,2}$} \\
    \large{$^1$ Dept. of ECE \& ASRI, $^2$ IPAI, Seoul National University, Seoul, Korea} \\
    {\tt\small \{cheeun914, kyoungmu\}@snu.ac.kr} \\
    \vspace{1.0em}
}

In this supplementary material, we present additional experimental results in \Cref{sup:exp}, additional analyses in \Cref{sup:analysis}, additional ablation study in \Cref{sup:hyperparam}, and additional qualitative results in \Cref{sup:qual}.

\section{Additional experiments}
\label{sup:exp}

\subsection{Comparison on scale 2}
In addition to the evaluations done in the main manuscript on SR networks of scale 4, we extend our evaluation to SR networks of scale 2.
First, we compare our method with existing adaptive quantization methods for SR in \Cref{tab:sup-scale2}.
For a fair comparison, we apply quantization to the body module following previous methods.
As shown in \Cref{tab:sup-scale2}, our method achieves a similar trade-off to existing adaptive quantization methods but with a significantly shorter process time.
We note that although PSNR/SSIM scores are lower on SRResNet ($\times2$), we incur lower computational costs (lower FAB). 
Furthermore, we compare our method with existing quantization methods without quantization-aware training (QAT) on scale 2 SR networks.
The results in \Cref{tab:sup-static-scale2} demonstrate that our method achieves competitive results against existing static quantization methods without QAT; our method results in lower computational complexity (FAB) and higher accuracy (PSNR/SSIM).

\begin{table*}[!ht]
\renewcommand{\arraystretch}{1.2}
\centering
\aboverulesep=0ex
\belowrulesep=0ex
\resizebox{\textwidth}{!}{
    \begin{tabular}{l|cc|c| c| cc cc cc}
        \toprule
        \multirow{2}{*}{Model} & \multirow{2}{*}{QAT} & \multirow{2}{*}{GT} & {Process} & \multirow{2}{*}{W / A} & 
        \multicolumn{2}{c}{Urban100} & \multicolumn{2}{c}{Test2K} & \multicolumn{2}{c}{Test4K}\\
        \cmidrule(lr){6-7} \cmidrule(lr){8-9} \cmidrule(lr){10-11}
        & & & Time & & FAB & PSNR / SSIM & FAB & PSNR / SSIM & FAB & PSNR / SSIM \\
        \midrule
        EDSR ($\times2$) & - & - & - & 32 / 32 & 32.0 & 31.98 / 0.927 & 32.0 & 32.76 / 0.928 & 32.0 & 34.37 / 0.944 \\
        \midrule
        EDSR-CADyQ & \cm & \cm & 47 hrs & 8 / 6MP & 6.1 & 31.90 / 0.927 & 5.8 & 32.70 / 0.928 & 5.7 & 34.31 / 0.943 \\
        EDSR-CABM  & \cm & \cm & 82 hrs & 8 / 6MP & 5.8 & 31.89 / 0.927 & 5.4 & 32.72 / 0.927 & 5.4 & 34.33 / 0.943 \\
        EDSR-AdaBM (\textbf{Ours}) & \xm & \xm & \textbf{103 sec} & 8 / 6MP & \textbf{5.8} & \textbf{31.86 / 0.927} & \textbf{5.5} & \textbf{32.73 / 0.928} & \textbf{5.4} & \textbf{34.33 / 0.943} \\
        \midrule
        SRResNet ($\times2$)& - & - & - & 32 / 32 & 32.0 & 31.60 / 0.923 & 32.0 & 32.60 / 0.927 & 32.0 & 34.20 / 0.942 \\
        \midrule
        SRResNet-CADyQ & \cm & \cm &  51 hrs & 8 / 6MP & 6.5 & 31.53 / 0.922 & 6.5 & 32.55 / 0.925 & 5.4 & 34.16 / 0.942 \\
        SRResNet-CABM  & \cm & \cm &  89 hrs & 8 / 6MP & 5.8 & 31.52 / 0.922 & 5.5 & 32.55 / 0.925 & 5.4 & 34.16 / 0.942 \\
        SRResNet-AdaBM (\textbf{Ours}) & \xm & \xm & \textbf{123 sec} & 8 / 6MP & \textbf{5.6} & \textbf{31.32 / 0.920} & \textbf{5.2} & \textbf{32.42 / 0.922} & \textbf{5.2} & \textbf{33.96 / 0.937} \\
        \bottomrule
    \end{tabular}
}
\vspace{-0.2cm}
\caption{
\textbf{Comparisons with adaptive quantization methods on SR networks of scale 2.}
}
\label{tab:sup-scale2}
\end{table*}

\begin{table*}[!ht]
\centering
\aboverulesep=0ex
\belowrulesep=0ex
\renewcommand{\arraystretch}{1.2}
\resizebox{\textwidth}{!}{
    \begin{tabular}{l|c|c| cc cc cc cc}
        \toprule
        \multirow{2}{*}{Model} & \multirow{2}{*}{FT} & \multirow{2}{*}{W / A} & 
        \multicolumn{2}{c}{Set5} & \multicolumn{2}{c}{Set14} & \multicolumn{2}{c}{BSD100} & \multicolumn{2}{c}{Urban100} \\
        \cmidrule(lr){4-5} \cmidrule(lr){6-7} \cmidrule(lr){8-9} \cmidrule(lr){10-11}& 
         && FAB & PSNR / SSIM & FAB & PSNR / SSIM & FAB & PSNR / SSIM & FAB & PSNR / SSIM \\
        \midrule 
        EDSR ($\times2$)&-&32 / 32 & 32.0 & 37.99 / 0.961 & 32.0 & 33.57 / 0.917 & 32.0 & 32.16 / 0.900 & 32.0 & 31.98 / 0.927\\
        \midrule
        EDSR-MinMax        &\xm& 4 / 4 & 4.0 & 32.87 / 0.850 & 4.0 & 30.48 / 0.818 & 4.0 & 29.55 / 0.799 & 4.0 & 28.92 / 0.821\\
        EDSR-Percentile    &\xm& 4 / 4 & 4.0 & 25.83 / 0.876 & 4.0 & 26.55 / 0.867 & 4.0 & 27.09 / 0.862 & 4.0 & 24.18 / 0.842\\
        EDSR-MinMax+FT     &\cm& 4 / 4 & 4.0 & 34.55 / 0.907 & 4.0 & 31.51 / 0.867 & 4.0 & 30.50 / 0.867 & 4.0 & 29.19 / 0.847\\
        EDSR-Percentile+FT &\cm& 4 / 4 & 4.0 & 29.69 / 0.915 & 4.0 & 28.77 / 0.884 & 4.0 & 28.86 / 0.876 & 4.0 & 26.23 / 0.864\\
        EDSR-PTQ4SR        &\cm& 4 / 4 & 4.0 & 36.88 / 0.947 & 4.0 & 32.81 / 0.904 & 4.0 & 31.59 / 0.886 & 4.0 & 30.60 / 0.907\\
        EDSR-AdaBM (\textbf{Ours}) &\cm& 4 / 4MP & \textbf{3.6} & \textbf{37.10} / \textbf{0.955} & \textbf{3.6} & \textbf{32.85} / \textbf{0.910} & \textbf{3.5} & \textbf{31.63} / \textbf{0.891} & \textbf{3.8} & \textbf{30.48} / \textbf{0.912} \\
        \midrule 
        RDN ($\times2$) &-&32 / 32 & 32.0 & 38.05 / 0.961 & 32.0 & 33.59 / 0.918 & 32.0 & 32.20 / 0.900 & 32.0 & 32.12 / 0.929\\
        \midrule
        RDN-MinMax        &\xm& 4 / 4 & 4.0 & 24.44 / 0.549 & 4.0 & 23.16 / 0.525 & 4.0 & 23.29 / 0.527 & 4.0 & 22.38 / 0.549 \\
        RDN-Percentile    &\xm& 4 / 4 & 4.0 & 23.33 / 0.918 & 4.0 & 23.39 / 0.757 & 4.0 & 24.86 / 0.859 & 4.0 & 21.47 / 0.848 \\
        RDN-MinMax+FT     &\cm& 4 / 4 & 4.0 & 33.63 / 0.930 & 4.0 & 30.53 / 0.878 & 4.0 & 29.76 / 0.856 & 4.0 & 27.13 / 0.851 \\
        RDN-Percentile+FT &\cm& 4 / 4 & 4.0 & 27.64 / 0.928 & 4.0 & 27.11 / 0.878 & 4.0 & 27.42 / 0.861 & 4.0 & 24.36 / 0.853 \\
        RDN-PTQ4SR        &\cm& 4 / 4 & 4.0 & 33.68 / 0.933 & 4.0 & 30.73 / 0.868 & 4.0 & 29.92 / 0.848 & 4.0 & 27.52 / 0.844 \\
        RDN-AdaBM (\textbf{Ours}) &\cm& 4 / 4MP & \textbf{3.8} & \textbf{34.90} / \textbf{0.932} & \textbf{3.7} & \textbf{31.42} / \textbf{0.885} & \textbf{3.6} & \textbf{30.37} / \textbf{0.863} & \textbf{3.8} & \textbf{28.34} / \textbf{0.864}\\
        \bottomrule
    \end{tabular}
}
\vspace{-0.2cm}
\caption{ 
\textbf{Comparisons with static quantization methods without QAT on SR networks of scale 2.}
}
\label{tab:sup-static-scale2}
\end{table*}

\subsection{Full quantization v.s. partial quantization}
In this work, we fully quantize SR networks to compare with existing static quantization methods without QAT.
However, most quantization methods on SR adopt partial quantization for the SR networks by only applying quantization to the body module of the network.
Thus, we analyze the effect of fully quantizing the network in \Cref{tab:sup-fq}.
Although partial quantization provides limited benefits in terms of cost reduction (\ie, the overall computational cost for the network remains larger), it results in higher reconstruction accuracy.
Overall, our method achieves higher accuracy with a lower computational cost in both partial and full quantization settings.
\begin{table*}[!ht]
\centering
\aboverulesep=0ex
\belowrulesep=0ex
\renewcommand{\arraystretch}{1.2}
\resizebox{\textwidth}{!}{
    \begin{tabular}{l|c|c| cc cc cc cc}
        \toprule
        \multirow{2}{*}{Model} & \multirow{2}{*}{FQ} & \multirow{2}{*}{W / A} & 
        \multicolumn{2}{c}{Set5} & \multicolumn{2}{c}{Set14} & \multicolumn{2}{c}{BSD100} & \multicolumn{2}{c}{Urban100} \\
        \cmidrule(lr){4-5} \cmidrule(lr){6-7} \cmidrule(lr){8-9} \cmidrule(lr){10-11}& 
         && FAB & PSNR / SSIM & FAB & PSNR / SSIM & FAB & PSNR / SSIM & FAB & PSNR / SSIM \\
        \midrule 
        EDSR &-&32 / 32 & 32.0 & 32.10 / 0.894 & 32.0 & 28.58 / 0.781 & 32.0 & 27.56 / 0.736 & 32.0 & 26.04 / 0.785\\
        \midrule
        EDSR-MinMax+FT     &\xm& 4 / 4 & 4.0 & 30.10 / 0.821 & 4.0 & 27.37 / 0.722 & 4.0 & 26.67 / 0.679 & 4.0 & 24.56 / 0.698\\
        EDSR-Percentile+FT &\xm& 4 / 4 & 4.0 & 31.15 / 0.876 & 4.0 & 27.96 / 0.769 & 4.0 & 27.21 / 0.727 & 4.0 & 25.12 / 0.757\\
        EDSR-PTQ4SR        &\xm& 4 / 4 & 4.0 & 31.23 / 0.864 & 4.0 & 28.02 / 0.757 & 4.0 & 27.17 / 0.713 & 4.0 & 25.28 / 0.746\\
        EDSR-AdaBM (\textbf{Ours}) &\xm& 4 / 4MP & \textbf{4.0} & \textbf{31.43} / \textbf{0.875} & \textbf{3.8} & \textbf{28.17} / \textbf{0.764} & \textbf{3.7} & \textbf{27.20} / \textbf{0.717} & \textbf{3.9} & \textbf{25.46} / \textbf{0.757} \\
        \midrule
        EDSR-MinMax+FT     &\cm& 4 / 4 & 4.0 & 28.97 / 0.821 & 4.0 & 26.47 / 0.721 & 4.0 & 26.24 / 0.687 & 4.0 & 23.46 / 0.674\\
        EDSR-Percentile+FT &\cm& 4 / 4 & 4.0 & 27.01 / 0.819 & 4.0 & 25.71 / 0.736 & 4.0 & 25.69 / 0.707 & 4.0 & 23.18 / 0.707\\
        EDSR-PTQ4SR        &\cm& 4 / 4 & 4.0 & 30.51 / 0.836 & 4.0 & 27.62 / 0.735 & 4.0 & 26.88 / 0.693 & 4.0 & 24.92 / 0.721\\
        EDSR-AdaBM (\textbf{Ours}) &\cm& 4 / 4MP & \textbf{3.8} & \textbf{31.02} / \textbf{0.860} & \textbf{3.7} & \textbf{27.87} / \textbf{0.751} & \textbf{3.5} & \textbf{26.91} / \textbf{0.700} & \textbf{3.7} & \textbf{25.11} / \textbf{0.736} \\
        \bottomrule
    \end{tabular}
}
\vspace{-0.2cm}
\caption{ \textbf{Comparisons between fully quantized networks and partially quantized networks.}
FQ denotes full quantization and the evaluation is done on EDSR of scale 4 that consists of 16 residual blocks (64 channels). 
}
\label{tab:sup-fq}
\end{table*}

\subsection{Comparison on CARN}
In addition to the networks evaluated in the main manuscript, we present an evaluation on CARN, a more lightweight SR model.
We compare our method with existing adaptive quantization methods for SR in \Cref{tab:sup-adaptive}.
The results indicate that AdaBM achieves a similar trade-off with existing methods, while the processing time is substantially accelerated to the second level.
Although CARN-AdaBM utilizes a higher average bit-width (FAB) compared to existing methods, it leads to improved reconstruction accuracy.
Moreover, we compare our method with static quantization methods without QAT in \Cref{tab:sup-static}.
Our adaptive method consistently outperforms existing methods with a lower FAB.
\begin{table*}[!ht]
\centering
\renewcommand{\arraystretch}{1.2}
\aboverulesep=0ex
\belowrulesep=0ex
\resizebox{\textwidth}{!}{
    \begin{tabular}{l|cc|c| c| cc cc cc}
        \toprule
        \multirow{2}{*}{Model} & \multirow{2}{*}{QAT} & \multirow{2}{*}{GT} & {Process} & \multirow{2}{*}{W / A} & 
        \multicolumn{2}{c}{Urban100} & \multicolumn{2}{c}{Test2K} & \multicolumn{2}{c}{Test4K}\\
        \cmidrule(lr){6-7} \cmidrule(lr){8-9} \cmidrule(lr){10-11}
        & & & Time & & FAB & PSNR / SSIM & FAB & PSNR / SSIM & FAB & PSNR / SSIM \\
        \midrule
        CARN & - & - & - & 32 / 32 & 32.0 & 26.07 / 0.784 & 32.0 & 27.70 / 0.782 & 32.0 & 28.77 / 0.814 \\
        \midrule
        CARN-CADyQ & \cm & \cm & 23 hrs & 8 / 6MP & 5.2 & 25.90 / 0.780 & 4.5 & 27.64 / 0.781 & 4.5 & 28.72 / 0.812 \\
        CARN-CABM  & \cm & \cm & 41 hrs & 8 / 6MP & 4.4 & 25.83 / 0.778 & 4.2 & 27.60 / 0.780 & 4.2 & 28.67 / 0.811 \\
        CARN-AdaBM (\textbf{Ours}) & \xm & \xm & \textbf{49 sec} & 8 / 6MP & \textbf{5.6} & \textbf{25.98 / 0.781} & \textbf{5.3} & \textbf{27.68 / 0.781} & \textbf{5.2} & \textbf{28.77 / 0.813} \\
        \bottomrule
    \end{tabular}
}
\vspace{-0.2cm}
\caption{ 
\textbf{Comparisons with adaptive quantization methods on CARN ($\times4$).}
}
\label{tab:sup-adaptive}
\end{table*}

\begin{table*}[!ht]
\renewcommand{\arraystretch}{1.2}
\centering
\aboverulesep=0ex
\belowrulesep=0ex
\resizebox{0.98\textwidth}{!}{
    \begin{tabular}{l|c|c| cc cc cc cc}
        \toprule
        \multirow{2}{*}{Model} & \multirow{2}{*}{FT} & \multirow{2}{*}{W / A} & 
        \multicolumn{2}{c}{Set5} & \multicolumn{2}{c}{Set14} & \multicolumn{2}{c}{BSD100} & \multicolumn{2}{c}{Urban100} \\
        \cmidrule(lr){4-5} \cmidrule(lr){6-7} \cmidrule(lr){8-9} \cmidrule(lr){10-11}& 
         && FAB & PSNR / SSIM & FAB & PSNR / SSIM & FAB & PSNR / SSIM & FAB & PSNR / SSIM \\
        \midrule 
        CARN ($\times4$) &-&32 / 32 & 32.0 & 32.14 / 0.893 & 32.0 & 28.61 / 0.781 & 32.0 & 27.58 / 0.736 & 32.0 & 26.07 / 0.784\\
        \midrule
        CARN-MinMax        &\xm& 4 / 4 & 4.0 & 30.94 / 0.874 & 4.0 & 27.82 / 0.760 & 4.0 & 27.01 / 0.715 & 4.0 & 25.06 / 0.749\\
        CARN-Percentile    &\xm& 4 / 4 & 4.0 & 26.55 / 0.806 & 4.0 & 25.75 / 0.729 & 4.0 & 25.78 / 0.696 & 4.0 & 23.42 / 0.703\\
        CARN-MinMax+FT     &\cm& 4 / 4 & 4.0 & 31.36 / 0.881 & 4.0 & 28.01 / 0.766 & 4.0 & 27.21 / 0.723 & 4.0 & 25.15 / 0.753\\
        CARN-Percentile+FT &\cm& 4 / 4 & 4.0 & 30.75 / 0.870 & 4.0 & 27.73 / 0.759 & 4.0 & 26.95 / 0.715 & 4.0 & 24.67 / 0.733\\
        CARN-PTQ4SR        &\cm& 4 / 4 & 4.0 & 31.41 / 0.881 & 4.0 & 28.03 / 0.766 & 4.0 & 27.19 / 0.722 & 4.0 & 25.22 / 0.755\\
        CARN-AdaBM (\textbf{Ours}) &\cm& 4 / 4MP & \textbf{3.7} & \textbf{31.68} / \textbf{0.885} & \textbf{3.6} & \textbf{28.23} / \textbf{0.771} & \textbf{3.4} & \textbf{27.30} / \textbf{0.726} & \textbf{3.6} & \textbf{25.45} / \textbf{0.762} \\
        \bottomrule
    \end{tabular}
}
\vspace{-0.2cm}
\caption{ 
\textbf{Comparisons with static quantization methods without QAT on CARN ($\times4$).}
}
\label{tab:sup-static}
\end{table*}

\section{Analysis}
\label{sup:analysis}
\subsection{Data sampling}
To obtain the calibration data, we randomly sampled data with a fixed random seed for our main manuscript experiments.
However, we found that different random seeds for data sampling yield different performances of the quantized model.
Here, we investigate the different sampling schemes for building the calibration dataset.
For example, we implement a stratified sampling scheme based on image complexity.
Images are divided into $N$ sub-groups based on the image gradient.
Then, random sampling is done for each sub-group.
As shown in \Cref{tab:sup-sampling}, such sampling gives additional gain but at the cost of additional processing time from forming the sub-groups.

\begin{table}[!ht]
    \renewcommand{\arraystretch}{1.2}
    \centering
    \aboverulesep=0ex
    \belowrulesep=0ex
    \resizebox{0.7\textwidth}{!}{
        \begin{tabular}{c|cccc}
            \toprule
            Sampling Method & FAB$_\downarrow$ & PSNR$_\uparrow$ & SSIM$_\uparrow$ & Processing Time \\
            \midrule
            Random & 3.80 $\pm$ 0.12 & 30.79 $\pm$ 0.21 & 0.857 $\pm$ 0.004 & 76 sec \\
            Stratified ($N$=4) & 3.68 & 30.80 & 0.853 & 85 sec\\
            Stratified ($N$=8) & 3.78 & 30.94 & 0.858 & 86 sec\\
            \bottomrule
        \end{tabular}
    }
    \vspace{-0.2cm}
    \caption{\textbf{Sampling methods} for 4-bit EDSR ($\times$4) on Set5. For random sampling, we average the result of different seeds.}
    \label{tab:sup-sampling}
\end{table}

\subsection{On-device latency}
Along with the speedup of time to obtain the quantized network, our framework also achieves speedup in inference time.
In \Cref{tab:sup-latency}, we report the latency of our quantized model on x86 and ARM CPUs.
Since only INT4/8 bits are supported for acceleration on current existing inference libraries, we upcast intermediate bits to INT8.
The results show that our framework is beneficial in terms of inference time.
We anticipate further speedup gain via acceleration on intermediate bits.

\begin{table}[!ht]
    \renewcommand{\arraystretch}{1.2}
    \centering
    \aboverulesep=0ex
    \belowrulesep=0ex
    \resizebox{0.6\textwidth}{!}{
    \begin{tabular}{l|lll}
        \toprule
        Method & EDSR & EDSR-CADyQ & EDSR-\textbf{AdaBM} \\
        \midrule
        x86 CPU & 4.002 sec & 0.974 sec ($\times$4.108) & 0.742 sec ($\times$5.391)\\
        ARM CPU & 3.998 sec & 1.880 sec ($\times$2.126) & 1.746 sec ($\times$2.290)\\
        \bottomrule
    \end{tabular}
    }
    \vspace{-0.2cm}
    \caption{\textbf{Average latency} for EDSR ($\times$4) on DIV2K validation set.}
    \label{tab:sup-latency}
\end{table}

\section{Ablations}
\label{sup:hyperparam}
We investigate the effect of hyperparameters used in our work: weight for bit loss ($\lambda_{bit}$), the percentile for calibrating the image-to-bit mapping module ($p_I$), and for calibrating the layer-to-bit mapping module ($p_L$).
As shown in \Cref{tab:sup-hyperparam-a}, the weight of bit loss controls the trade-off between accuracy and computational complexity.
Reducing the bit loss weight can cause the bit mapping modules to select overall higher bit-widths, prioritizing minimal reconstruction loss.
Consequently, a smaller $\lambda_{bit}$ results in higher PSNR/SSIM but uses more computational costs (\ie, larger FAB).
However, employing a large $\lambda_{bit}$ strictly restricts the average bit-width from increasing, resulting in a model with smaller computational cost but lower PSNR/SSIM.
Our framework can achieve varying levels of trade-off by controlling $\lambda_{bit}$, but we fix $\lambda_{bit}=50$ in our experiments.
Additionally, the results in \Cref{tab:sup-hyperparam-b} and \Cref{tab:sup-hyperparam-c} justify our choice of hyperparameters.
\begin{table}[!h]
\renewcommand{\arraystretch}{1.2}
\centering
\aboverulesep=0ex
\belowrulesep=0ex
    \subfloat[Ablation on $\lambda_{bit}$ \label{tab:sup-hyperparam-a}]{
        \centering
        \resizebox*{!}{0.1\textheight}{
            \begin{tabular}{c|cc}
                \toprule
                $\lambda_{bit}$ & FAB$_\downarrow$ & PSNR$_\uparrow$ / SSIM$_\uparrow$ \\
                \midrule
                1 & 4.78 & 31.38 / 0.872 \\
                10 & 4.08 & 31.10 / 0.865 \\
                \textbf{50} & {3.78} & {31.02} / {0.860}\\
                100 & 3.72 & 30.89 / 0.858 \\
                \bottomrule
            \end{tabular}
        }
    }
    \hspace{5mm}
    \subfloat[Ablation on $p_I$ \label{tab:sup-hyperparam-b}]{
        \centering
        \resizebox*{!}{0.1\textheight}{
            \begin{tabular}{c|cc}
                \toprule
                $p_I$ & FAB$_\downarrow$ & PSNR$_\uparrow$ / SSIM$_\uparrow$ \\
                \midrule
                5 & 3.99 & 31.13 / 0.864 \\
                \textbf{10} & {3.78} & {31.02} / {0.860}\\
                20 & 3.85 & 30.93 / 0.860 \\
                30 & 3.99 & 30.79 / 0.858 \\
                \bottomrule
            \end{tabular}
        }
    }
    \hspace{5mm}
    \subfloat[Ablation on $p_L$ \label{tab:sup-hyperparam-c}]{
        \centering
        \resizebox*{!}{0.1\textheight}{
            \begin{tabular}{c|cc}
                \toprule
                $p_L$ & FAB$_\downarrow$ & PSNR$_\uparrow$ / SSIM$_\uparrow$ \\
                \midrule
                5 & 3.84 & 31.00 / 0.858 \\
                10 & 3.96 & 31.06 / 0.860 \\
                20 & 3.84 & 31.05 / 0.860 \\
                \textbf{30} & {3.78} & {31.02} / {0.860}\\
                \bottomrule
            \end{tabular}
        }
    }
    \vspace{-2mm}
    \caption{ \textbf{Effect of hyperparameters} evaluated on Set5 with 4-bit EDSR ($\times$4). 
    }
    \label{tab:sup-hyperparam}
\end{table}

\clearpage

\section{Additional qualitative results}
\label{sup:qual}

\subsection{Qualitative comparison}
\label{sup:comp}
\begin{figure*}[!h]
\centering
\begin{center}
    \setlength{\tabcolsep}{0.9mm}
    \newcommand{\w}{0.18\linewidth}
    \begin{tabular}{ccccc}
        \centering
        \includegraphics[width=\w]{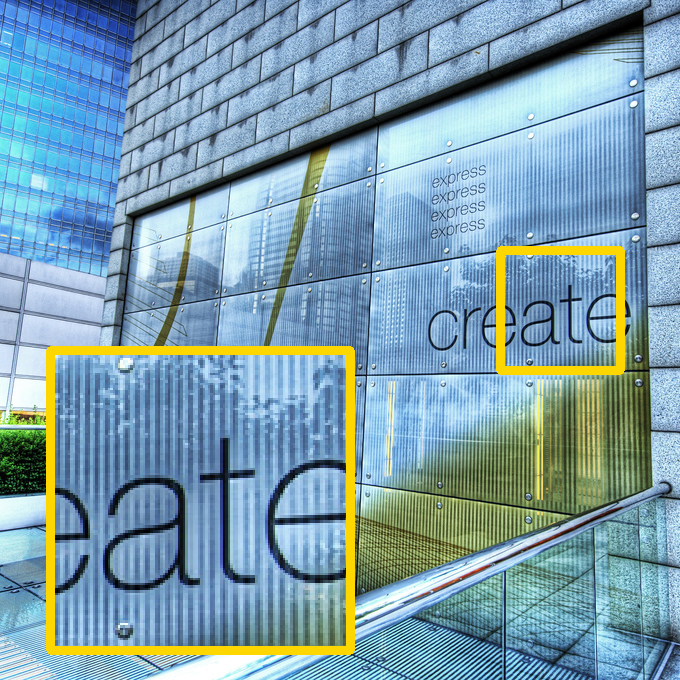} & \includegraphics[width=\w]{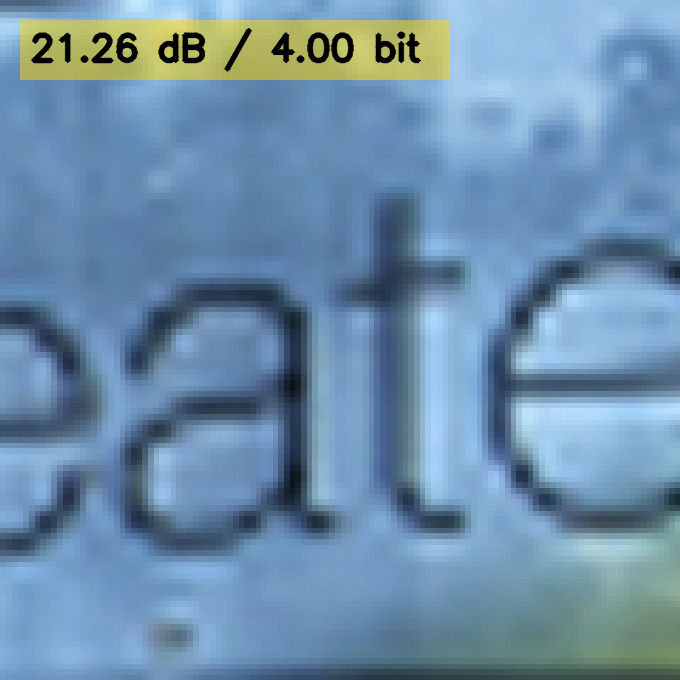} & \includegraphics[width=\w]{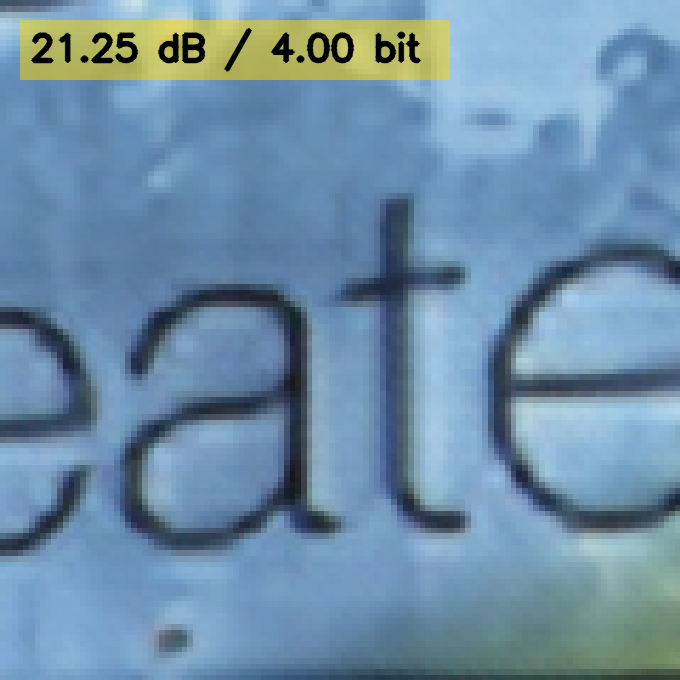} & \includegraphics[width=\w]{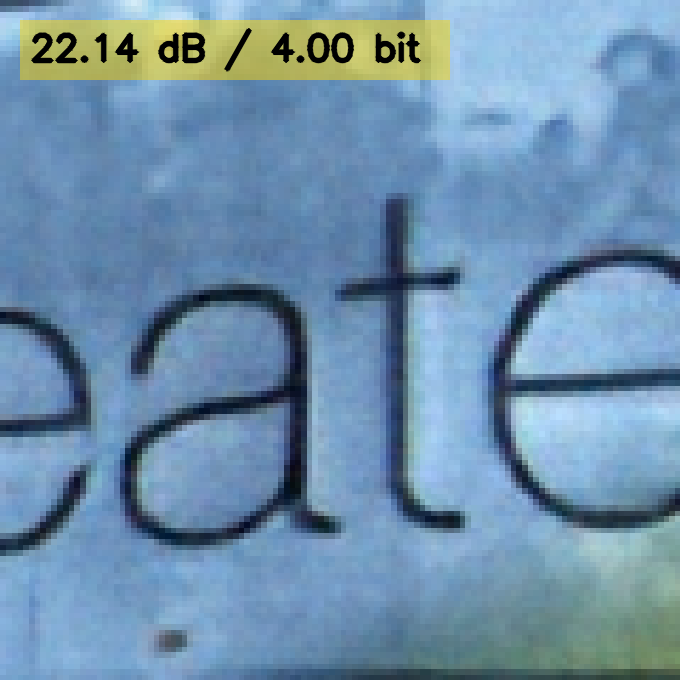} & \includegraphics[width=\w]{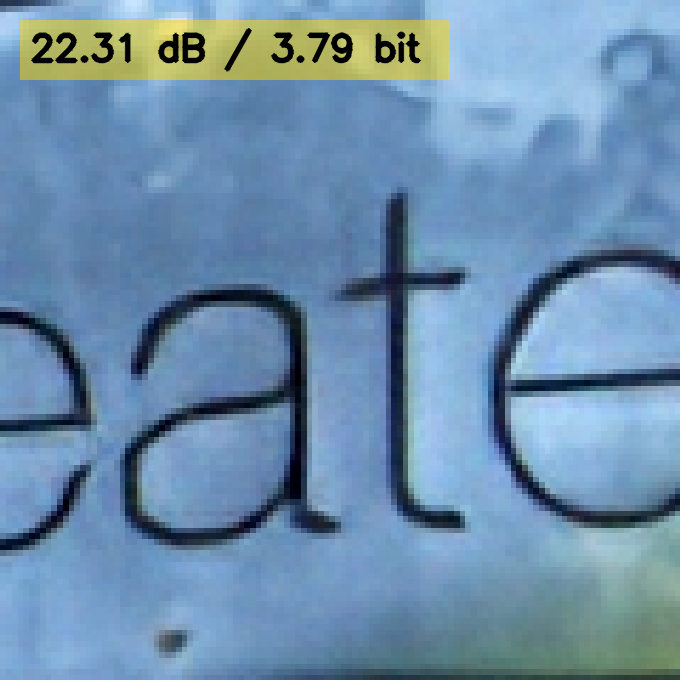} \\
        \small GT (img060) & \small EDSR-MinMax+FT & \small EDSR-Percentile+FT &\small EDSR-PTQ4SR &\small \textbf{EDSR-AdaBM} \vspace{2mm}\\
        \includegraphics[width=\w]{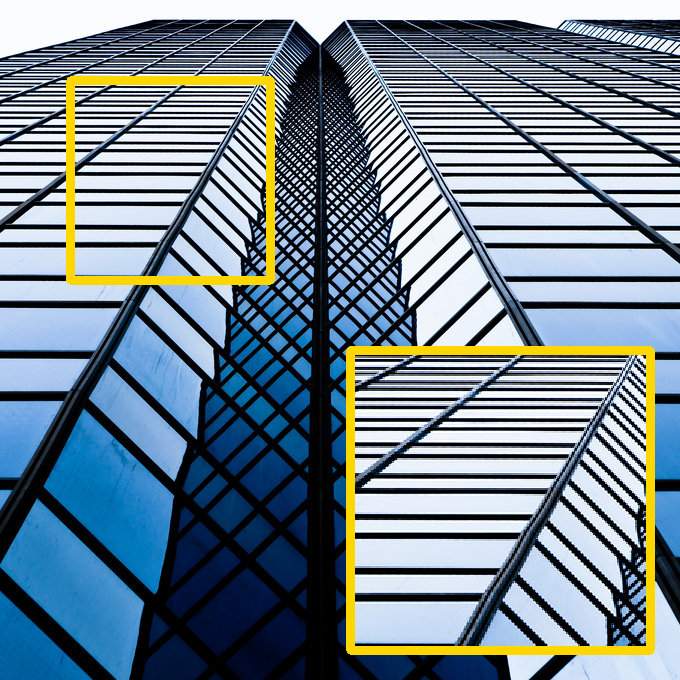} & \includegraphics[width=\w]{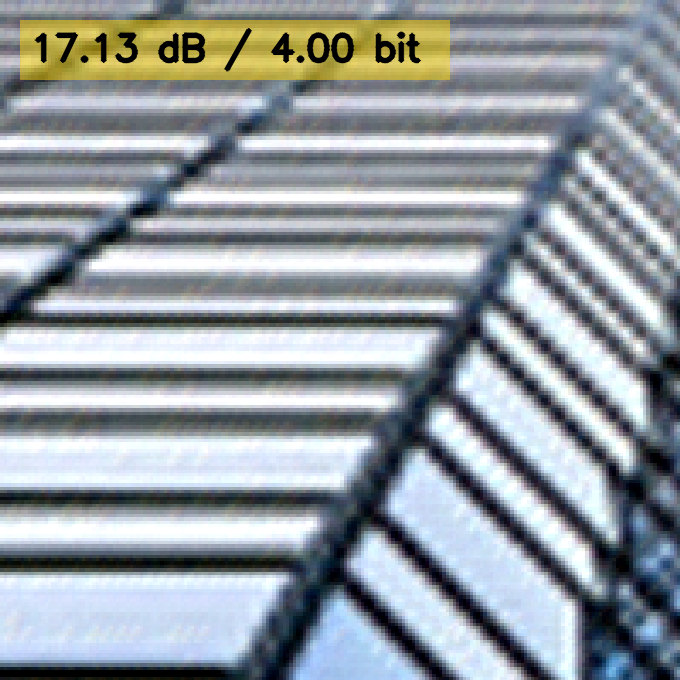} & \includegraphics[width=\w]{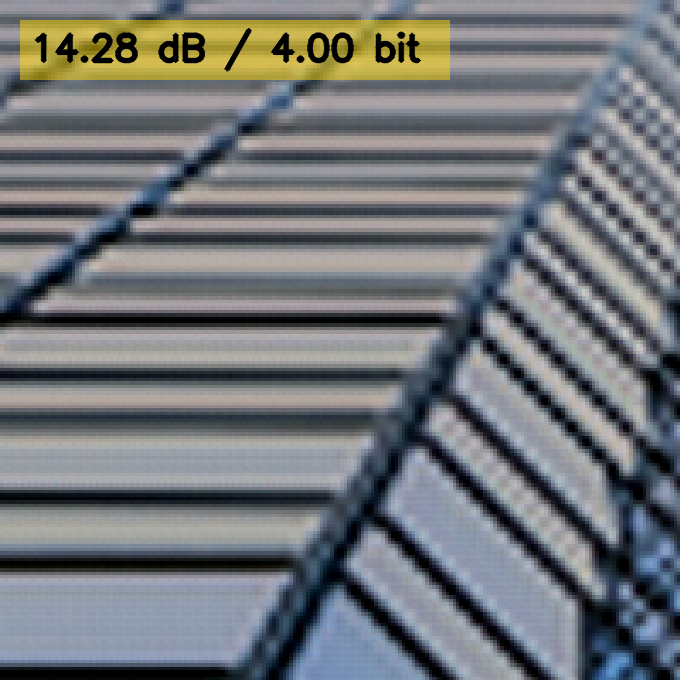} & \includegraphics[width=\w]{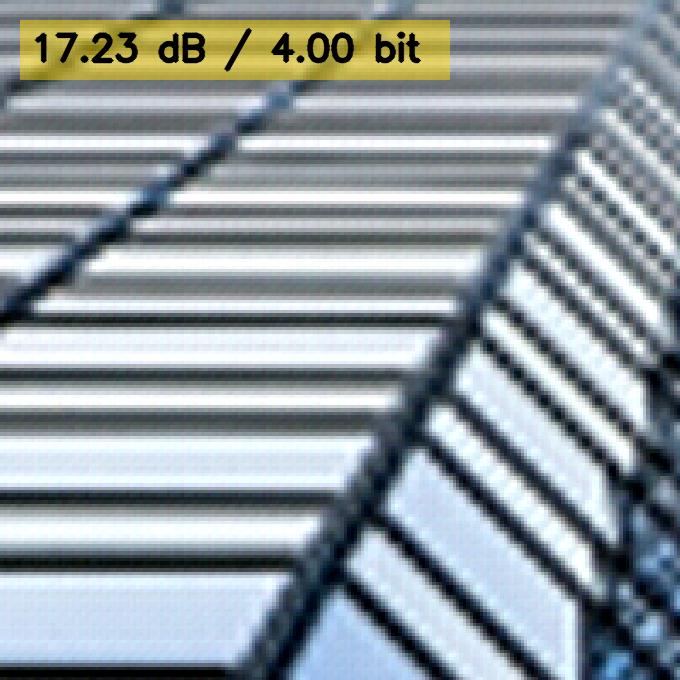} & \includegraphics[width=\w]{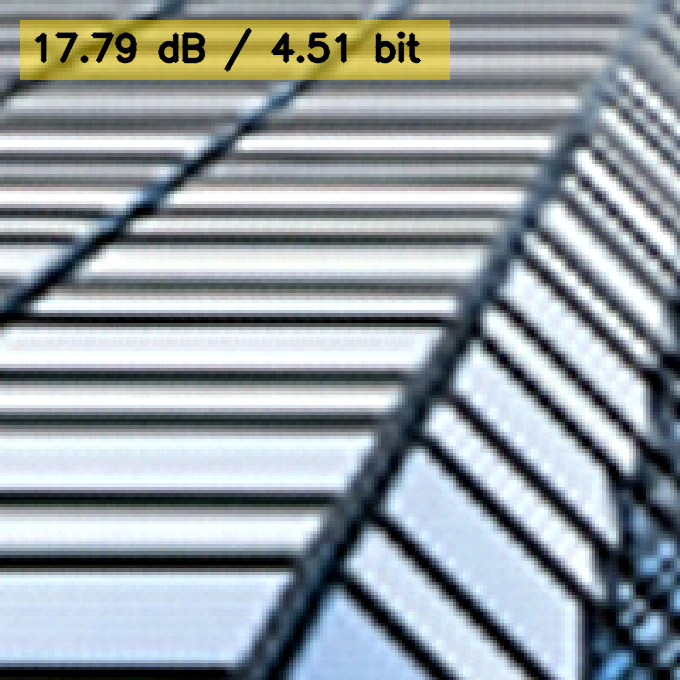} \\
        \small GT (img067) & \small RDN-MinMax+FT & \small RDN-Percentile+FT &\small RDN-PTQ4SR &\small \textbf{RDN-AdaBM} \vspace{2mm} \\
    \end{tabular}
    \vspace{-0.2cm}
    \caption{
    \textbf{Qualitative results} on 4-bit SR networks of scale 4. The networks are fully quantized.
    }
    \vspace{-0.3cm}
    \label{fig:sup-qual}
\end{center}
\end{figure*}

\subsection{Visualization}
\label{sup:vis}
\begin{figure*}[!h]
    \centering
    \begin{minipage}[b]{0.22\linewidth}
        \centering
        \includegraphics[width=\textwidth]{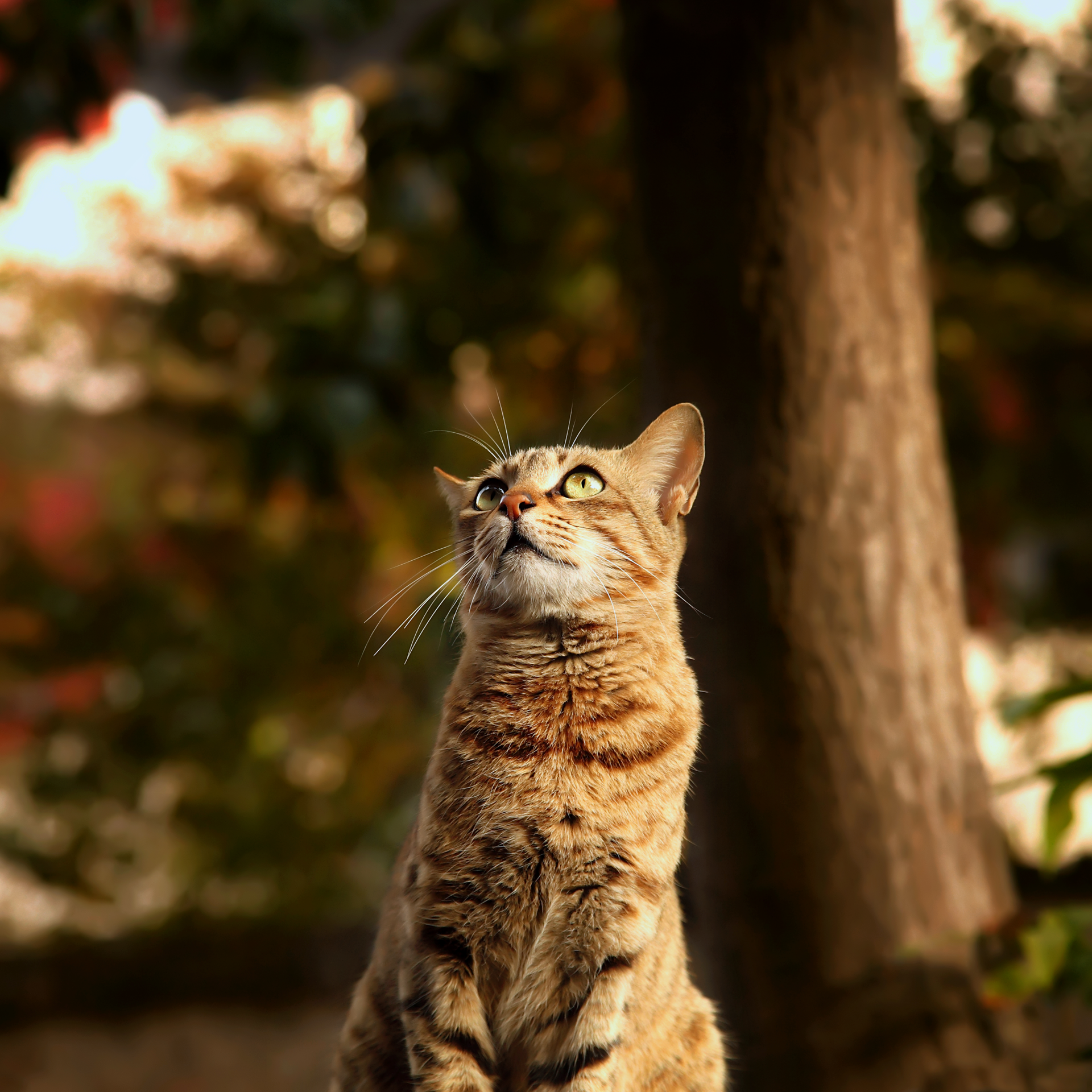}
    \end{minipage}
    \hspace{1mm}
    \begin{minipage}[b]{0.22\linewidth}
        \centering
        \includegraphics[width=\textwidth]{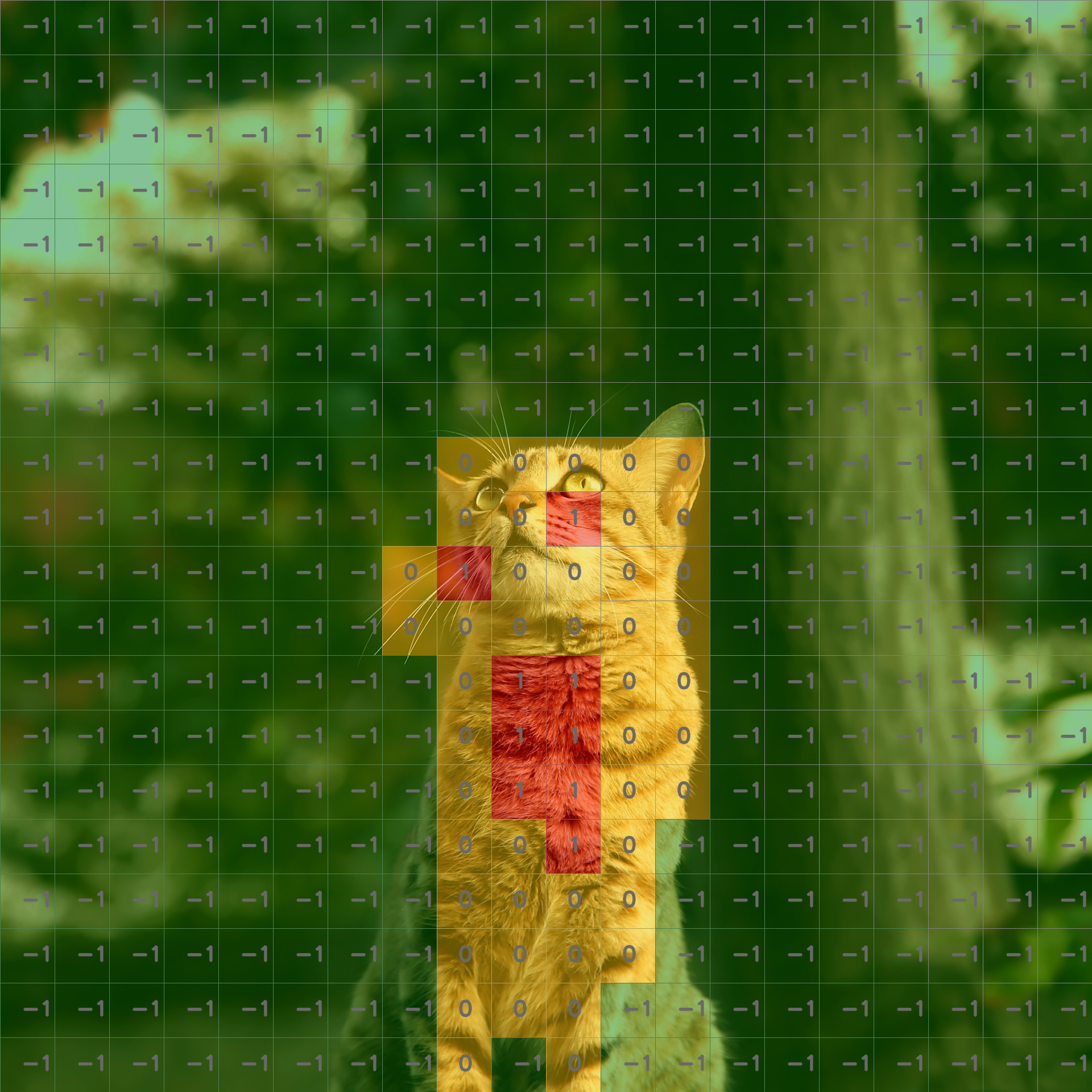}
    \end{minipage}
    \hspace{-5mm}
    \begin{minipage}[b]{0.5\linewidth}
        \centering
        \includegraphics[height=0.2\textwidth]{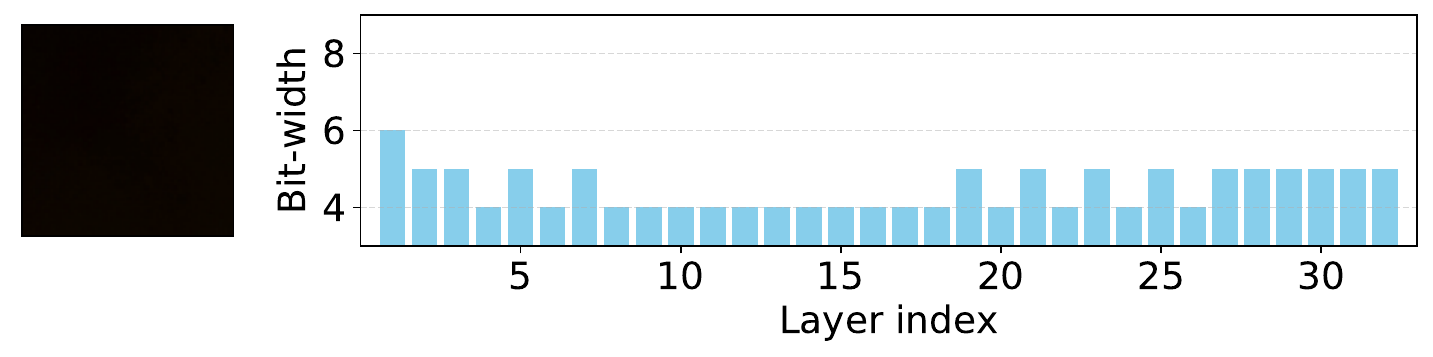} 
        \includegraphics[height=0.2\textwidth]{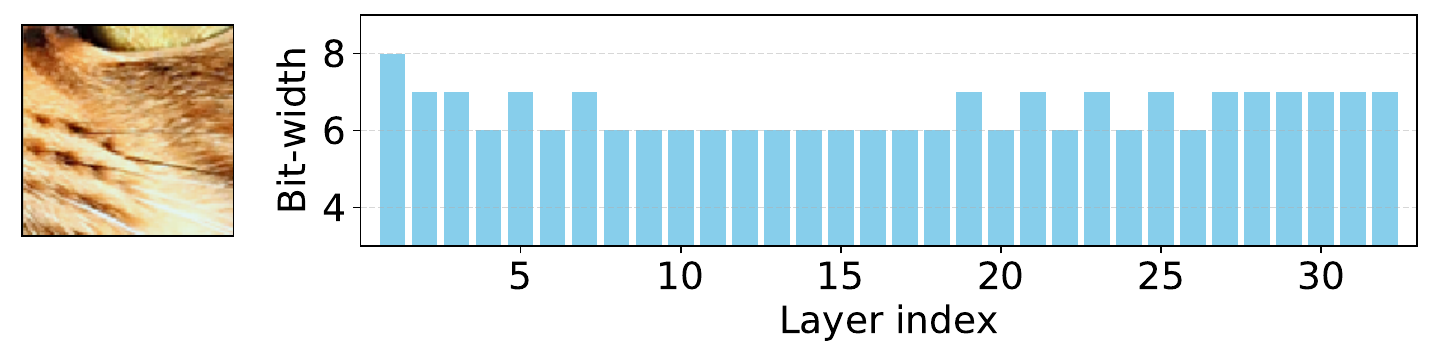}
    \end{minipage}
    \vspace{2mm} \\
    (a) EDSR-AdaBM \\
    \vspace{2mm}
    \begin{minipage}[b]{0.22\linewidth}
        \centering
        \includegraphics[width=\textwidth]{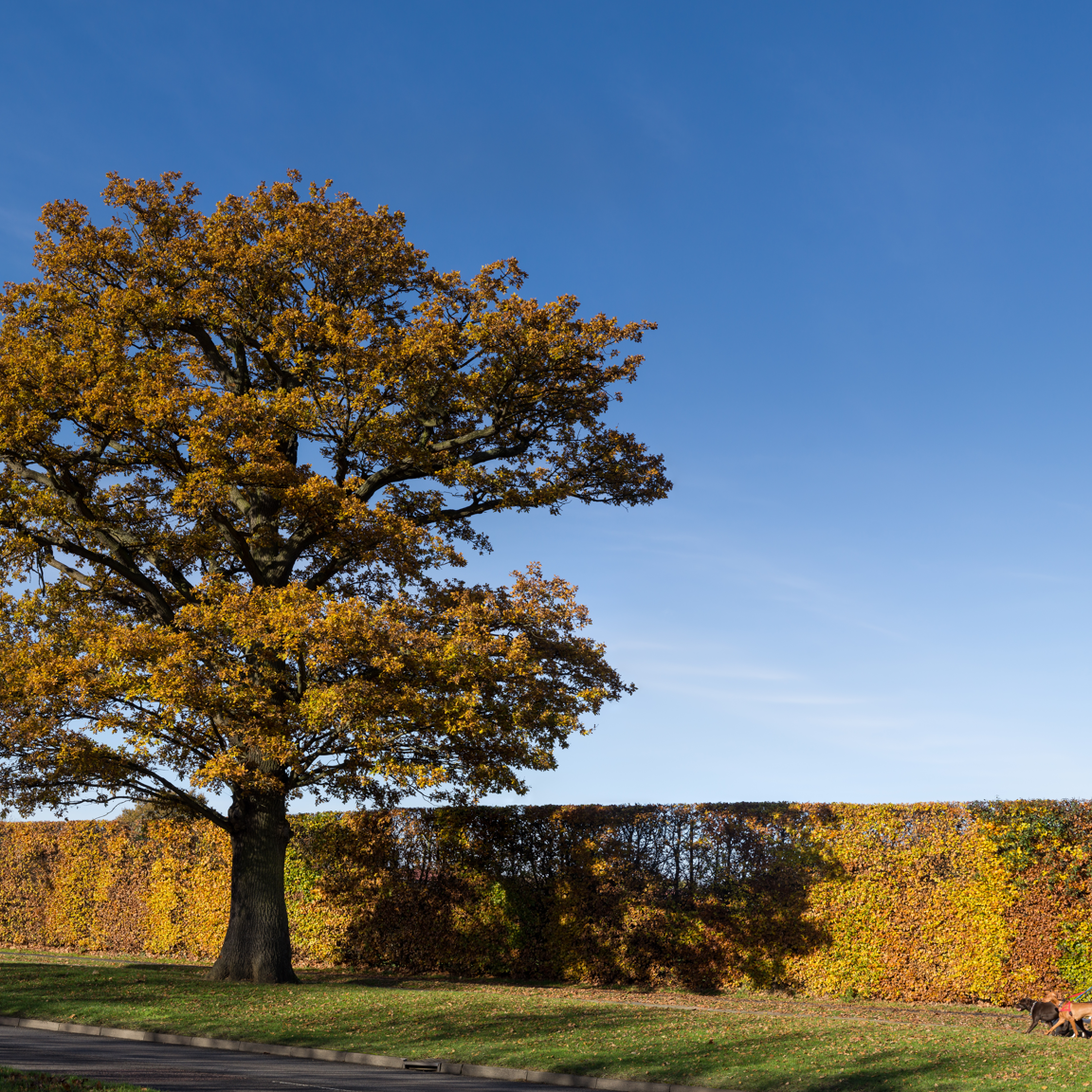}
    \end{minipage}
    \hspace{1mm}
    \begin{minipage}[b]{0.22\linewidth}
        \centering
        \includegraphics[width=\textwidth]{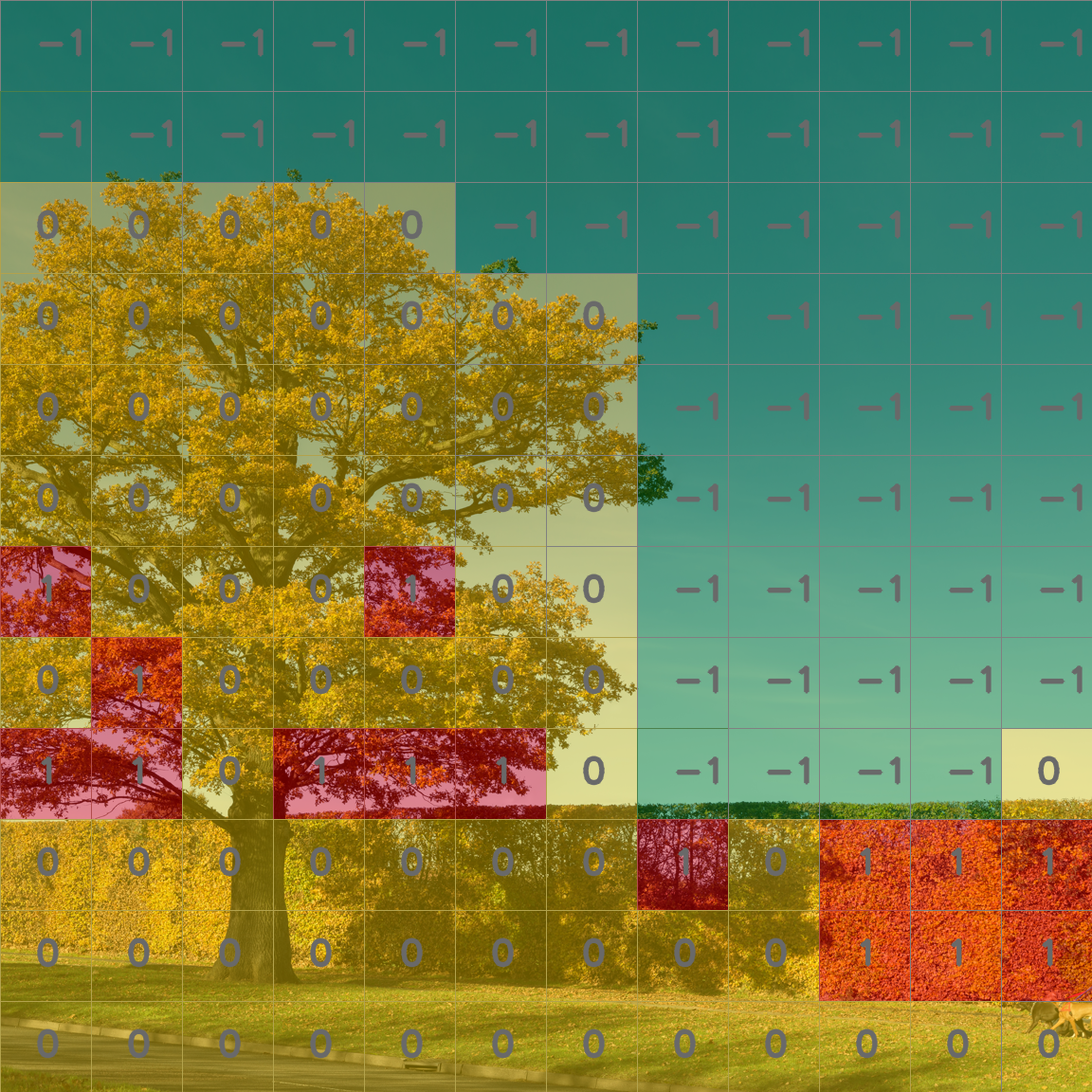}
    \end{minipage}
    \hspace{-5mm}
    \begin{minipage}[b]{0.5\linewidth}
        \centering
        \includegraphics[height=0.2\textwidth]{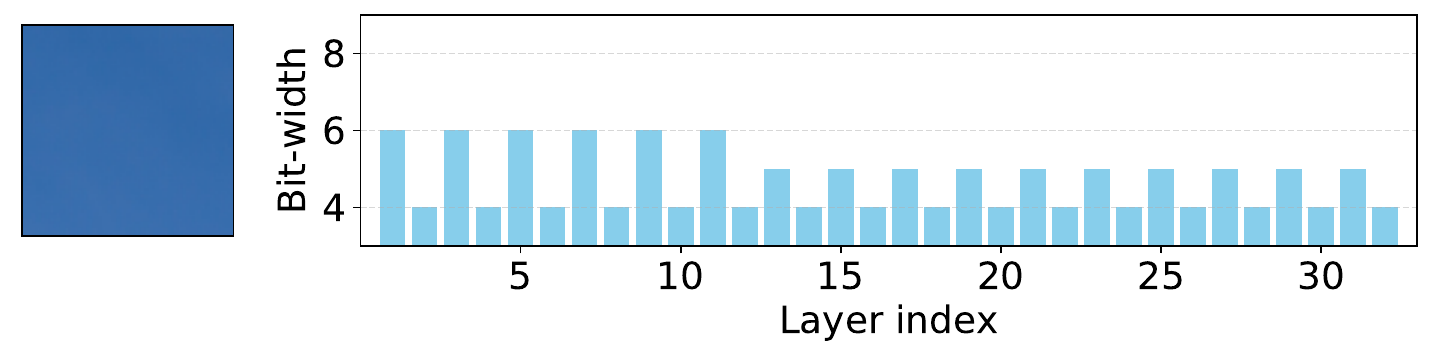} 
        \includegraphics[height=0.2\textwidth]{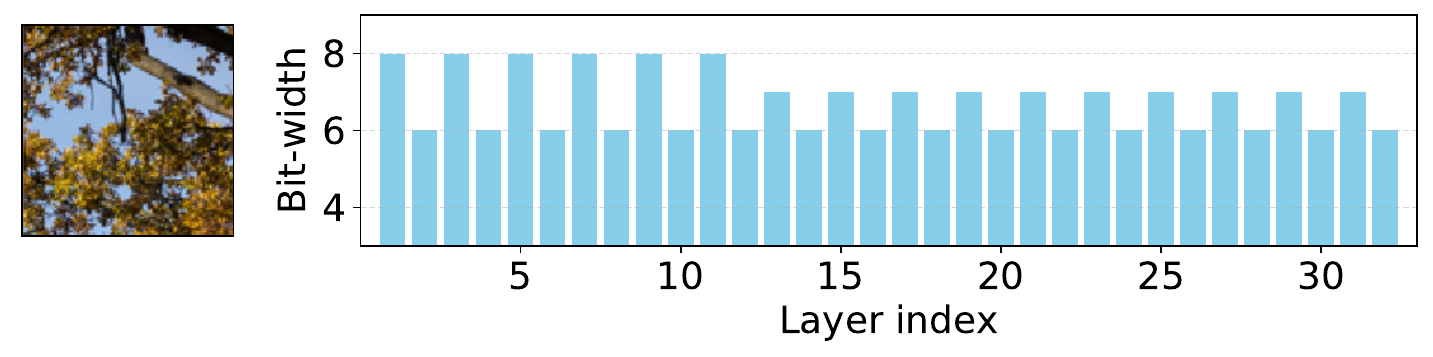}
    \end{minipage}
    \vspace{2mm} \\
    (b) SRResNet-AdaBM \\
    \vspace{2mm}
    \caption{
        \textbf{Visualization of adaptive bit-mapping of AdaBM} on large inputs. Evaluation done on SR networks of scale 4. 
    }
\label{fig:sup-vis}
\end{figure*}

\end{document}